\documentclass{article}

\usepackage{microtype}
\usepackage{graphicx}
\usepackage{subfigure}
\usepackage{booktabs} % for professional tables

\usepackage{hyperref}

\usepackage[accepted]{icml2023}

\usepackage{amsmath}
\usepackage{amssymb}
\usepackage{mathtools}
\usepackage{amsthm}

\usepackage[capitalize,noabbrev]{cleveref}

\theoremstyle{plain}
\newtheorem{theorem}{Theorem}[section]
\newtheorem{proposition}[theorem]{Proposition}

\newtheorem{definition}[theorem]{Definition}

\newtheorem{example}[theorem]{Example}
\theoremstyle{remark}

\newcommand{\allatomrules}{\Tilde{\mathcal{C}}}
\newcommand{\atomrule}{c}
\newcommand{\vecallrules}{\mathbf{\Tilde{r}}}
\newcommand{\vecallfiredrules}{\mathbf{r}}
\newcommand{\randomvecallrules}{\mathbf{\Tilde{R}}}
\newcommand{\randomvecfiredrules}{\mathbf{R}}
\newcommand{\oneruleassignment}{\Tilde{r}}

\newcommand{\onerulRV}{\Tilde{R}}
\newcommand{\oneFiredrulRV}{R}
\newcommand{\triple}{t}
\newcommand{\alltriples}{\mathcal{G}}

\newcommand{\indexSetAll}{\Tilde{I}}
\newcommand{\indexSetFired}{I}
\newcommand{\tFiredAtomRulesG}{\mathcal{C}_\triple(\alltriples)}
\newcommand{\probloglogicprogram}{\mathcal{P}}

\usepackage{times}
\usepackage{soul}
\usepackage{url}
\usepackage[utf8]{inputenc}
\usepackage[small]{caption}
\usepackage{graphicx}
\usepackage{amsmath}
\usepackage{amsthm}
\usepackage{amssymb}
\usepackage{booktabs}
\usepackage{algorithm}
\usepackage{algorithmic}
\usepackage{rotating}
\usepackage{listings}
\usepackage{graphicx}
\usepackage{color}
\usepackage{amssymb}
\usepackage{amsmath}
\usepackage{booktabs}
\usepackage{multirow}
\usepackage{appendix}
\usepackage{pgfplots}
\usepackage{adjustbox}
\usepackage{tabularx}
\usepackage{hyperref}
\usepackage{float}

\usepackage{bm}
\usepackage{spverbatim}
\usepackage[textsize=tiny]{todonotes}

\icmltitlerunning{On the Aggregation of Rules for Knowledge Graph Completion}

\begin{document}

\twocolumn[
\icmltitle{On the Aggregation of Rules for Knowledge Graph Completion}




\begin{icmlauthorlist}
\icmlauthor{Patrick Betz}{yyy}
\icmlauthor{Stefan Lüdtke}{xxx}
\icmlauthor{Christian Meilicke}{yyy}
\icmlauthor{Heiner Stuckenschmidt}{yyy}
\end{icmlauthorlist}

\icmlaffiliation{yyy}{University of Mannheim, Germany}
\icmlaffiliation{xxx}{University of Rostock, Germany}

\icmlcorrespondingauthor{Patrick Betz}{patrick@informatik.uni-mannheim.de}

\icmlkeywords{Machine Learning, Knowledge Graph Completion, Rule Learning, Rule Aggregation}

\vskip 0.3in
]



\printAffiliationsAndNotice{}  % leave blank if no need to mention equal contribution

\begin{abstract}
Rule learning approaches for knowledge graph completion are efficient, interpretable and  competitive to purely neural models. The rule aggregation problem is concerned with finding one plausibility score for a candidate fact which was simultaneously predicted by multiple rules. Although the problem is ubiquitous, as data-driven rule learning can result in noisy and large rule sets, it is underrepresented in the literature and its theoretical foundations have not been studied before in this context. In this work, we demonstrate that existing aggregation approaches can be expressed as  marginal inference operations over the predicting rules. In particular, we show that the common Max-aggregation strategy, which scores candidates based on the rule with the highest confidence, has a probabilistic interpretation. Finally, we propose an efficient and overlooked baseline which combines the previous strategies and is competitive to computationally more expensive approaches.
\end{abstract}

%$\mathdutchcal{r}_i \in \allatomrules$.........$r_i \in \allatomrules $
\section{Introduction}

A knowledge graph (KG) is a collection of \textit{relation(subject, object)} facts which can be used to compactly describe certain domains. KGs can be utilized for various downstream applications such as drug repurposing~\cite{hetioliu2021neural} or visual relationship detection~\cite{Baier2017ImprovingVR}. Most of the real-world KGs are incomplete, which means that absent facts are not necessarily false. The problem of knowledge graph completion (KGC) aims to derive the missing facts by using the information in the existing graph~\cite{ruffinelli2020dog,rossi2020knowledge}. The proposed model classes in the literature are data-driven, e.g., a model might learn the regularity that people which appear in movies tend to be actors and can use it to make new predictions. Although the dominating paradigm in the literature lies on models based on latent representation, a KG is symbolic by its nature.

Symbolic machine learning approaches for KGC employ rule mining techniques and represent the KG with the raw predicates which makes them inherently interpretable. In regard to predictive performance they are shown to be competitive to latent based approaches~\cite{rossi2020knowledge} and can achieve state-of-the-art results on large graphs~\cite{meilicke2023largeAnyburl}. To perform KGC with a symbolic approach, a previously learned set of rules has to be applied to the KG to derive plausibility scores for unseen target facts. Whenever multiple rules predict a candidate fact, the question arises of how to aggregate individual rules, as demonstrated in the following running example.
%\begin{RunningExp}
\begin{example} \label{ex:running}
Consider the following clauses or rules.

\vspace{-0.5cm}
\resizebox{\linewidth}{!}{
\begin{minipage}{\linewidth}
\begin{align*}
      \textrm{$\atomrule$}_1 \; [0.64]: \; &\textrm{wf(X,Y)} \leftarrow \textrm{internAt(X,Y)}   \\
      \textrm{$\atomrule$}_2 \; [0.44]: \; &\textrm{wf(X,Y)} \leftarrow \textrm{studentAt(X,A), locIn(A,B), locIn(Y,B)}  \\
      \textrm{$\atomrule$}_3 \; [0.41]: \;  &\textrm{wf(X,Y)} \leftarrow \textrm{studentAt(X,A), cooperatesWith(A,Y)}
\end{align*}
\end{minipage}
}

Here $\textrm{wf}$ represents the relation $\textrm{worksFor}$ and $\textrm{locIn}$ represents $\textrm{locatedIn}$. The numbers in brackets denote rule confidences, i.e., the proportion of correct predictions on a training KG. The first and third rule are quite intuitive. The second rule expresses that a person might work for a company if that company is located at the same place where this person went to university. Now assume that all three rules predict \textit{Anna} to work for \textit{Google}. The rule aggregation problem is concerned with finding a final score derived from the three confidence values. The aggregation will also reflect if, e.g., \textit{Anna} is more likely to work at \textit{Google} than a person for which only the first two rules made the prediction. 
\end{example}

While combining logical reasoning and probabilistic uncertainty is a fundamental aspect of statistical relational learning~\cite{muggleton1996stochastic,kersting2001towards,richardson2006markov}, the aggregation problem is often not expressed explicitly. Additionally, these approaches perform model theoretic entailment, which is too expensive in our settings, as KGs can consist of a large number of facts with millions of learned rules. Similarly, in the field of association rule mining, rule quality is often estimated for individual rules independently without considering the problem of  aggregation~\cite{amie2013,ChenWG16,OrtonaMP18,fan2022discovering}.

The predictive quality of a mined rule set depends to a large extend on the aggregation decision and surprisingly there exists a theoretical and empirical gap in the recent KGC literature between techniques to learn rules and their successful application. To the best of our knowledge, there only exist two recent works which are primarily concerned with the aggregation problem for KGC \cite{ott2021safran,betz2022supervised}. While they improve upon simple strategies, the approaches are computationally expensive and theoretical foundations are not discussed.

The goal of this work is to close this gap and to inspire new research in this direction. We aim to achieve this by developing the formal foundations of the problem and by empirically analysing the practicality of existing approaches. We present a probabilistic model in which the aggregation reduces to performing marginal inference over a joint distribution of the rules when rule marginals are approximated with confidences (Section~\ref{sec: representation} and~\ref{sec: inference over the knowledge graph}). With this formulation we are able to show that the common Max-aggregation strategy can be recovered from the model when the correlation matrix of the rules is set to the upper Fréchet-Hoeffding bound for the correlation of random variables (Section~\ref{sec: sub probab-rule-agg}). We then search for the simplest and most efficient way to combine the assumptions made by common aggregation strategies. This leads to an efficient baseline, Noisy-or top-$h$, which  is competitive when taking into account the performance-runtime trade-off (Section~\ref{sec:experiments}). Moreover, our experiments show that the choice of the aggregation function has significant performance impacts and therefore it deserves more attention in the context of rule-based KGC.

\section{Related Work}
While data-driven rule learning approaches for KGC are often evaluated in comparison to embedding models, the focus of this work is rule aggregation and we therefore refer to the recent literature for an overview to latent-based KGC~\cite{rossi2020knowledge}.

Rule mining approaches learn datalog rules from a KG. In the context of \textbf{association rule mining}, AMIE~\cite{amie2013} and the respective improved versions AMIE+~\cite{amieplus2015} and AMIE3~\cite{lajus2020fast} show how to mine rules when data is incomplete. AnyBURL~\cite{meilicke2019anyburl} is the successor of RuleN~\cite{rulen2018}. It is shown to be competitive to neural approaches~\cite{rossi2020knowledge,meilicke2023largeAnyburl} and it can be utilized to explain predictions made by embedding models~\cite{betz2022adversarial}. Other approaches are tailored towards large graphs~\cite{fan2022discovering,ChenWG16} or to learn negative rules~\cite{OrtonaMP18}.
There also exist attempts to improve rule quality by providing more advanced confidence computations~\cite{amie2013,pellissier2017completeness,zupanc2018estimating}. The rule quality is evaluated by calculating the precision of the individual rules independent from the remaining rules on a gold standard KG. For the resulting metrics, the aggregation problem is irrelevant. In this work we regard rule quality from the viewpoint of the predictions made by the rules, which also allows comparisons to other model classes. 

Related branches of work combine latent and symbolic models in hybrid approaches~\cite{guo2016jointly,guo2018knowledge,DBLP:conf/uai/Garcia-DuranN18,wu2022learning,meilicke2021naive}. Moreover, some work propose \textbf{differentiable rule learning} i.e., learning rules by solving a smooth optimization problem~\cite{yang2017differentiable,sadeghian2019drum}. Rule mining and the aggregation are arguably coalesced in one forward pass of a neural module. It has been shown, nevertheless, that the rules extracted from the models might not derive the same facts as the models themselves and achieve a lower predictive performance \cite{faithful22Cucala}. Therefore, they might benefit from encapsulating rule learning and the aggregation. A step in this direction is made by RNNlogic~\cite{qu2020rnnlogic}, in which a neural rule generator and a reasoning predictor operate independently. The predictive performance of the resulting model, when not augmented with embeddings, lacks, however, in regard to purely symbolic models.

The combination of logic and uncertainty has a rich history in the \textbf{statistical relational learning} literature. For instance, Stochastic Logic Programs~\cite{muggleton1996stochastic,sato1997prism} and Bayesian Logic Programs (BLP)~\cite{kersting2001towards} augment inductive logic programming~\cite{muggleton1994inductive} with probability semantics. Rules are represented as conditional probabilities and a joint probability distribution is modelled over the least Herbrand base of the logic program. Here, the aggregation problem becomes explicit. In particular, when multiple conditionals have the same effect variable, they are collapsed into one by the use of a \textit{combining rule}. Nevertheless, this heuristic is applied on top of the formal framework whereas in this work we model the problem directly. A difficulty for BLPs is that the probability distribution is only well defined when the underlying graph does not contain cycles which is quite unlikely in the context of KGC when millions of rules are learned. Markov Logic Networks (MLNs)~\cite{richardson2006markov} are proposed to overcome the cycle problem as well as the requirement to define the \textit{ad hoc} combining rule. MLNs subsume many of the approaches from the statistical learning literature. Each possible ground fact is associated with a binary random variable and every possible grounding of every rule with a weight and a binary feature.
The aggregation of clauses is performed implicit for MLNs and can not be modelled easily. We show an example regarding MLNs in the appendix of this work.

The focus of this work are settings where model theoretic entailment is not feasible. For instance, an MLN would need to define $15k \cdot 237$ random variables on the dataset FB15k-237~\cite{toutanova2015observed} and a feature for every possible rule grounding with a ruleset size of 5 million. Even if we  would just calculate the immediate predictions of the rules on this dataset, including storing some indices for further processing, this would already take more than 600GB of memory. A similar note can be made for neural theorem proofing, where the forward-chaining algorithm is relaxed to a smooth differentiable function~\cite{evans2018learning,rocktaschel2017endNTP1,minervini2020differentiableGNTP,minervini2020learningCNTP}. To the best of our knowledge, these approaches have not shown yet to scale to datasets of the size used in our experiments. This also holds for ProbLog~\cite{de2007problog} which combines probabilistic inference with model theoretic entailment and has the strongest resemblance to our approach. We discuss the details in Section~\ref{sec: sub probab-rule-agg} and in the appendix of this work.

The rule aggregation problem is discussed explicitly by SAFRAN~\cite{ott2021safran} where a clustering of the rules is learned and by Betz et al.~\yrcite{betz2022supervised} who represent rules with embeddings. These works show improvements in regard to simple strategies but they do not consider a fundamental treatment of the problem and the models are inefficient to use, which will be demonstrated in the experimental section.

\section{Background} \label{sec: prelim}
%\textcolor{blue}{It is common in the context of rule aggregation in application scenarios such as knowledge graph completion \textcolor{red}{(abbrev)} to suppress the underlying logical and probabilistic representation, for instance, rules/clauses are interpreted solely in a computational manner by what they can predict when grounding their rule bodies with a knowledge base. While this can be seen as an noncomplete approach it is also necessary for computational and conceptual efficiency \textcolor{red}{(explain this)}. We will therefore introduce the concepts in the viewpoint of their respective fields and then gradually in the remaining sections close the gap to probabilistic modelling and reasoning.}

\subsection{Knowledge Graph Completion}\label{sec: KGC}
A KG $\alltriples$ is a set of \textit{relation(subject, object)} triples or facts with $\alltriples \subseteq \mathcal{E} \times \mathcal{P} \times \mathcal{E}$ where $\mathcal{E}$ denotes a set of entities and $\mathcal{P}$ a set of binary predicates which we term relations. KGC is concerned with finding unknown facts, given an input or training KG $\alltriples$. In this work, we focus on the mostly used evaluation protocols which are defined by ranking based evaluation metrics. The derivations of this work are, however, independent of the evaluation protocol as long as scalar scores for candidate predictions are required.

The common practice is to split the graph into disjoint training, validation, and testing sets. %$\alltriples^{train}, \alltriples^{valid}, \alltriples^{test}$ with $\alltriples^{train} \cap \alltriples^{valid} \cap, \alltriples^{test} = \emptyset$.
After the training or mining phase a model is evaluated by proposing answers to queries formed from the facts in the test set. For each of these evaluation facts a head query and a tail query are formed. For example, from $worksFor(Anna,Google)$ the queries  $worksFor(Anna, ?)$ and $worksFor(?, Google)$ are formed, where $worksFor$ is a relation and $Anna$ and $Google$ are entities. A model has to propose candidate facts for the tail query, e.g., $worksFor(Anna, e_1)$ and candidate facts for the head query $worksFor(e_2, Google)$ for multiple $e_1, e_2 \in \mathcal{E}$.  Each candidate fact is assigned with a score such that for each direction a ranking of answers can be formed. The metrics usually are presented with their filtered versions, e.g., if $e_2 \neq Anna$ but $worksFor(e_2, Google)$ exists in one of the data splits, then it is removed from from the ranking of the current query to not penalize the model when it correctly ranks true answers on top positions. Performance is measured by the ranking position of the respective true candidate $worksFor(Anna,Google)$ in both directions where the mean reciprocal rank (MRR) and Hits@X being the most common evaluation metrics. The definitions of the metrics can be found in the appendix.

\subsection{Rules and Application} \label{sec: Application}
We let a $c \in \allatomrules$ denote a logical clause, which we will term rule throughout the work, where $\allatomrules$ is a collection of clauses. The $c$ will later be indexed and represented by separate random variables. The rules that we consider in this work are of the form as given in the running example. They are composed of variables and relations and they additionally can contain entities as shown in the following example.
\begin{align*}
    speaks(X,English)\leftarrow livesIn(X,London)
\end{align*}
We call $speaks(X,English)$ the head of the rule and $livesIn(X,London)$ the body of the rule. The rules and the KG can be described with a subset of Prolog, where entities are constants, relations are predicates, rules are clauses, and the facts of the KG are ground atoms where we do not consider negation. We will use the rule learners AnyBURL~\cite{meilicke2019anyburl} and AMIE3~\cite{lajus2020fast} in our experimental section and we refer to the respective works for further details, nevertheless, the descriptions and derivations in this work are independent of the particular syntax.

We define a substitution to be the expression obtained when replacing the variables of the rules with entities from $\mathcal{E}$. For instance, for the first rule from the running example with $(X{=}Anna, Y{=}Google)$ we obtain the substitution $worksFor(Anna,Google) {\leftarrow} internsAt(Anna, Google)$. A detailed formalization is suppressed here for brevity.

Rule application refers to predicting previously unseen facts given a set of rules and the input or training KG. We can describe it compactly with the recently introduced concept of one-step-entailment~\cite{betz2022adversarial}. Let $\allatomrules$ be a set of rules and $\alltriples$ a KG.

\begin{definition}[One-step entailment $\models_1$]\label{def:one-step}
The fact $t$ is \textbf{one-step entailed} by $\: \allatomrules \cup \alltriples$, written as $\: \allatomrules \cup \alltriples \models_1 t$, iff there is a rule in $\allatomrules$ for which a substitution exists such that the resulting body facts are in $\alltriples$ and the head is equal to $t$.
\end{definition}
Clearly, one-step entailment is weaker but more efficient than model theoretic entailment. As mentioned before, we focus on settings where general entailment is not feasible. One-step-entailment implies entailment but not vice versa.\footnote{Note that $\models_1$ is different to $\bar{k}$-entailment which limits the number of constants used in entailment~\cite{kuzelka2018pac}.} In the context of KGC often the less formal notion of an individual rule predicting a candidate is used which we can now describe precisely.\footnote{A formalization with the immediate consequence operator in the logic programming context is likewise possible.} 
\begin{definition}[Prediction]\label{def:prediction}
A rule $\atomrule \in \allatomrules$ \textbf{predicts} a fact $t$ iff it individually one-step entails $t$, i.e., iff $\{\atomrule\} \cup \; \alltriples \models_1 t$.
\end{definition}
For simplicity, we will write $c \models_1 t$ instead of $\{\atomrule\} \cup \; \alltriples \models_1 t$, where from the context the reference to the facts $\alltriples$ will be clear.
%Finally, for a fact $\triple$ we let $\tFiredAtomRulesG = \{\atomrule \;|\; \atomrule \models_1 t \; \mathrm{ and } \; \atomrule \in \allatomrules \}$ denote the set of predicting rules with $\tFiredAtomRulesG \subseteq \allatomrules$.
The section concludes with an example.

\begin{example}[cont.] \label{ex:prediction-and-one-step}
Let $e_d$, $e_u$, and $e_g$ be entities in $\mathcal{E}$.
%Let $cw$, $ia$, $sa$ denote the relations, $cooperatesWith$, $internsAt$, $studentAt$, respectively.
Let $\triple = wf(e_d,e_g)$ and assume that
\begin{equation*}
\mathcal{G}=\left\{
  \begin{array}{@{}ll@{}ll@{}}
    & cooperatesWith(e_g, e_u) &  \\
    & internAt(e_d, e_g) &  \\
    & studentAt(e_d, e_u) &  
  \end{array}\right\}.
\end{equation*}
Consider the three rules from the running example. Then the joint set of rules and every pairwise set of rules one-step entail $\triple$ while only the first and the third rule predict $\triple$.

\end{example}
%-------------------------------------------------------
\subsection{Rule Aggregation} \label{sec: Aggregation}
For the remainder of the work we assume that $\allatomrules$ is a given rulset that has been learned from the training graph $\alltriples$. Furthermore, for a target triple $\triple \notin \alltriples$ we let $\tFiredAtomRulesG$ denote the set of rules that predicted $\triple$ with respect to the KG $\alltriples$.
For performing KGC under any evaluation protocol a model has to assign plausibility scores to candidate facts. For rule-based KGC this requires the introduction of two additional concepts, rule confidences and aggregation strategies. 

%From the viewpoint of symbolic approaches two concepts are required that exceed the scope of the logical formalism of section \ref{sec: foundations} and \ref{sec: Application}, that is, rule confidences and aggregation strategies. 

\subsubsection{Confidences}
Rule confidences originate from the context of association rule mining and we will now assume that each rule in $\allatomrules$ is assigned with a confidence which can be calculated as follows.
\begin{align}
    \textit{conf}(\atomrule) = \frac{\big|\{t' \; |\;  \atomrule \models_1 t' \land t' \in \alltriples \}\big|}{\big|\{t' \; |\;  \atomrule \models_1 t'\}\big|} \label{eq: rul-conf}
\end{align}
Equation~\eqref{eq: rul-conf} is the vanilla confidence definition described in many works~(e.g., Gal\'{a}rraga et al.,~\yrcite{amie2013}). The confidence divides the number of all true predictions a rule makes by the number of all predictions of the rule. Intuitively, we could interpret this as the probability that the rule is true, which will be discussed in later sections. %Note that when the data is split into training and evaluation sets the confidence must be estimated based on the training graph only. %\textcolor{red}{(check if this stays in after related work is done} There exist also slight modifications of the confidence computation~\cite{pellissier2017completeness,zupanc2018estimating} which is, however, not the focus of this work.

\subsubsection{Aggregation Strategies}
In practical scenarios it rarely occurs that a candidate fact is predicted by only one rule, i.e., then $|\tFiredAtomRulesG|>1$. The rule aggregation problem, also termed joint prediction~\cite{amieplus2015}, is concerned with defining a function that maps the confidences of the rules that predicted the candidate to a real valued score.

Note that the number of rules that predict a candidate fact simultaneously can be large, as mentioned before, such that rules are to some extend redundant. For instance, if the second rule from the running example predicts \textit{Anna} to work for \textit{Google}, the question arises whether the third rule provides additional evidence for this prediction. The rules make the prediction for seemingly similar reasons, as it is more likely for an university and a company to cooperate when they are located in the same location. In the following the two most common aggregation strategies are defined.

\begin{definition}[Max-Aggregation]
The Max-Aggregation score $s^{M}$ is calculated according to the rule with the highest confidence from the rules that predicted the candidate, $s^{M}(t) = \max \{\textit{conf}(\atomrule) \;|\; \atomrule \in \tFiredAtomRulesG  \} $.
\end{definition}

\noindent Max-aggregation was first used in the context of KGC by Gal\'{a}rraga et al.~\yrcite{amieplus2015} and it was later adapted to \textit{Max+ aggregation}~~\cite{meilicke2019anyburl} which allows for tie handling. When the two predicting rules with the highest confidences for two candidates are identical the candidates are compared according to the rules with the second highest confidence which is continued until the candidates can be discriminated. 

\begin{definition}[Noisy-or aggregation]
The Noisy-or score $s^{NO}$ is calculated as the noisy-or product over the predicting rules,  $s^{NO}(t) = 1 - \prod_{\atomrule \in\tFiredAtomRulesG } (1- conf(\atomrule))$.
\end{definition}

\noindent The Noisy-or product originates from Bayesian networks where it is used to express independent causes~\cite{pearl1988probabilistic} and it was proposed by Gal\'{a}rraga et al.~\yrcite{amieplus2015} for KGC.

\begin{example}[cont] \label{ex:aggregation example}
Let us assume that $Anna$ is predicted by all rules from the starting example to work for $Google$,  while $Lisa$ is predicted by only the second and third rule to work for $Google$. The Max-aggregation and Noisy-or scores for $Anna$ are $0.64$ and $0.88$, respectively. For $Lisa$ they are $0.44$ and $0.67$. 
\end{example}

While the aggregation functions have the purpose of merging the various confidences into a final score, this value also should be meaningful in the sense that a higher value for one prediction should mean it is more likely than another prediction.

\section{Probabilistic and Efficient Rule Aggregation}\label{sec: main probab aggregation}
%-------------------------------------------------------------------------

In the following section we present the notation for the probabilistic representation, subsequently we introduce the inference model and show how the introduced rule aggregation functions can be recovered from the framework when making certain dependency assumptions. Finally, we will present an efficient baseline, that combines these assumptions.

\subsection{Representation}\label{sec: representation}
First, we enumerate the rules in $\allatomrules$ with an index set $\indexSetAll = \{1, ... , N\}$ such that  $\atomrule_i \in \allatomrules$ for $i \in \indexSetAll$. Each rule $\atomrule_i$ is represented by a binary random variable $\onerulRV_i$ which is also indexed by $\indexSetAll$ and has realisations $\oneruleassignment_i \in \{0,1\}$. We let $\randomvecallrules$ denote the random vector representing all rules and likewise $\vecallrules = (\oneruleassignment_i)_{i \in \indexSetAll} \in \{0, 1\}^N$ is the vector of realisations. For brevity we write $p(\vecallrules)$ for $p(\randomvecallrules{=}\vecallrules)$, that is, the probability that $\randomvecallrules$ takes value $\vecallrules$.

For the rule aggregation problem the set of rules $\tFiredAtomRulesG \subseteq \allatomrules$ that predict a target fact $\triple$ based on $\alltriples$ are of particular relevance. Therefore, similar as above $\tFiredAtomRulesG$ is enumerated by $\indexSetFired=\{1, ..., k\}$ and the random vector $\randomvecfiredrules$ with realisations $\vecallfiredrules = (r_j)_{j \in I} \in \{0, 1\}^k$ represents the rules that predict the target. Note that $\randomvecfiredrules$ represents a subset of all the rules and this depends on $t$, however, to not clutter notation we not write this explicit and the reference to $\triple$ will be clear from the context. 

Moreover, we write $p_j$ or $p_i$ for the probability that a rule is true, i.e.,  for the marginals $p(\oneFiredrulRV_j{=}1)$ or $p(\onerulRV_i{=}1)$. We assume that index sets are ordered according to the marginals, e.g., $p_m \geq p_n$ when $m \leq n$ with $m,n$ being indices. Facts $t$ are likewise represented as binary variables, here we overload notation for brevity and write $p(t)$ for the probability of a query triple to be true. For an observed triple $t \in \alltriples$ we set $p(t)=1$.

\subsection{Dealing with Uncertainty}
To incorporate uncertainty into the prediction of new facts we take the following approach. If we are certain that a rule is true, then we deduce that a prediction it makes must be also true. We can model this for all the learned rules with a conditional distribution that conditions on the truth values of the rules and the data.
\begin{align}
   p(t |  \mathbf{\vecallrules}, \alltriples)  = \left\{ \begin{array}{l} \label{eq:logical part}
    1, \textrm{ if $L(\mathbf{\vecallrules}) \models_1 t$   } \ \\
    0, \textrm{ else},
  \end{array}\right.
\end{align}
Here, $L$ is a simple mapping that collects all rule objects in $\allatomrules$ whose realisation are one in $\vecallrules$ and takes the union with $\alltriples$, i.e.,
\begin{align}
    L_{{\hat{I}}}^{\alltriples}: \vecallrules \mapsto  L_{{\hat{I}}}^{\alltriples}(\mathbf{\vecallrules}) = \{\atomrule_i  \;| \; \oneruleassignment_i = 1 \; \textit{and} \; i \in \hat{I} \} \cup \alltriples.
\end{align}
We drop, as shown in equation~\eqref{eq:logical part}, the reference to the index set $\indexSetAll$ and $\alltriples$ from $L$ for readability. Clearly, if the rules would not be associated with uncertainty evaluating equation~\eqref{eq:logical part} would boil down to performing rule application in regard to the correct rules. However, the truth values of the rules cannot be observed from the data.

We have, on the other hand, an estimate that statistically quantifies the uncertainty of the rules, the defined rule confidences. A confidence may serve as an approximation for the marginal probability that the respective rule is true, i.e., $p(\onerulRV_i{=}1)$. However, we have to acknowledge that it is only the marginal $\sum p(\onerulRV_i=1, \vecallrules_{-i})$, which sums over all realisations of the remaining rules, where $\vecallrules_{-i}$ is the vector of realisations with $\oneruleassignment_i$ dropped.

The last paragraph makes the difference to the viewpoint of association rule mining explicit. In fact, we assume that $p(\onerulRV_i{=}1)$ is potentially influenced by an underlying joint distribution. For instance, the confidence of the rule $c_2$ of the running example might be influenced by the confidence of $c_3$ through the second term in the sum $p(\onerulRV_2{=}1) = p(\onerulRV_2{=}1, \onerulRV_3{=} 0) +  p(\onerulRV_2{=}1, \onerulRV_3{=} 1)$. Therefore, for fact prediction associated with uncertainty we have to take into account the joint distribution over the rules which will be discussed in the next section.

\subsection{Inference for Target Facts} \label{sec: inference over the knowledge graph}
We want to calculate the probability that an unknown target fact $t \notin \alltriples$ is true, given the known triples, i.e., we seek to compute  $ p(t | \alltriples)$. However, we cannot observe the truth values $\vecallrules$ of the rules from the data and we therefore choose a standard approach regarding such settings, i.e., we marginalize over all possible rule realisations, 
%longer version
%\begin{align}
%    p(t | \alltriples) &= \sum_{\vecallrules \in \{0, 1\}^N}  p(t, %\vecallrules | \alltriples) \\
%    &= \sum_{\vecallrules \in \{0, 1\}^N}  p(t |\vecallrules, \alltriples) %p(\vecallrules | \alltriples), \label{eq: probab model final}
%\end{align}
\begin{align}
    p(t | \alltriples) = \sum_{\vecallrules \in \{0, 1\}^N}  p(t |\vecallrules, \alltriples) p(\vecallrules | \alltriples). \label{eq: probab model final}
\end{align}
Where we set $p(t |\vecallrules, \alltriples)$ to equation~\eqref{eq:logical part}. We can simply calculate $p(t |\vecallrules, \alltriples)$ by collecting all rules that are one in $\vecallrules$ and subsequently evaluate if one of these rules predicts the target, i.e., performing rule application. The distribution $p(\vecallrules | \alltriples)$ seems to be more problematic. It defines the joint distribution over all $N$ rules, given the data, including the rules that did not predict $\triple$. Rule aggregation, however, was defined with only the $k$ rules that predicted a candidate. We will argue in the following proposition that under one-step entailment for calculating $p(t | \alltriples)$ it is indeed sufficient also under the probabilistic model to exclusively take into account the rules $\randomvecfiredrules$ with realisations $\vecallfiredrules$ that predicted $\triple$.

\begin{proposition} \label{prop: kvsN}
    Under a one-step entailment regime, i.e., using equation~\eqref{eq:logical part} for $p(t |\vecallrules, \alltriples)$, and a global distribution $p(\vecallrules | \alltriples)$ we have that
    \begin{align}
     p(t | \alltriples) = \sum_{\vecallfiredrules\in \{0, 1\}^k}  p(t |\vecallfiredrules, \alltriples) p(\vecallfiredrules | \alltriples). \label{eq: modelWithKrules}  
    \end{align}
\end{proposition}

\noindent The proof is in the appendix. Instead of using the global distribution we can focus directly on performing marginal inference $p(\vecallfiredrules | \alltriples)$ with respect to the rules that predicted $\triple$. Although marginal inference can equally be expensive, the complexity can be reduced if the joint distribution is specified accordingly and if some parameters of the joint are known such as the individual rule marginals. Additionally, it might even be beneficial to model $p(\vecallfiredrules | \alltriples)$ directly. %\textcolor{red}{(note that, well, the question here arises when is it actually cheaper to compute $p(\vecallfiredrules | \alltriples)$ as it is still the marginalization over all other rules? e.g. SPN has cheap marginal inference..., but would you not have to argue for whatever global distribution you choose that the marginal inference is efficient (e.g. max, noisy-or, max-group)?) } 

Note that Proposition~\eqref{prop: kvsN} would not hold if we would consider general model theoretic entailment. Finally, by the definition of equation~\eqref{eq:logical part} and one-step entailment it is easy to see that the query probability is the probability that at least one rule from $\randomvecfiredrules$ is true.

\begin{proposition} \label{prop: oneIsTrue}
    For the query probability it holds that
    \begin{align}
    p(t | \alltriples)=p\big( \sum_{j \in \indexSetFired} \oneFiredrulRV_j \geq 1 \;|\; \alltriples\big). 
    \end{align}
\end{proposition}
\textbf{Proof.} We write out $p(t |\vecallfiredrules, \alltriples)$ in equation~\eqref{eq: modelWithKrules} and then drop the one term that is zero. The proposition follows from the definition of one-step-entailment as $L(\vecallfiredrules)$ one-step entails the target if at least one component of $\vecallfiredrules$ is one. That means the probabilities of all realisations where at least one rule is true are summed up. \qed

We will henceforth refer to calculating $p(\triple | \alltriples)$ under the previous derivations when mentioning the inference model and we conclude the section with an example.

\begin{example}[cont] \label{ex:only-predicting rules example}
Lisa is predicted by the two rules $\atomrule_2$ and $\atomrule_3$ to work for Google. Assuming that we know the joint distribution over all rules, we can calculate the probability that Lisa works for Google by querying the joint distribution for the probability that at least one of $\atomrule_2$ and $\atomrule_3$ is true. 
\end{example}

%-------------------------------------------------------

\subsection{Recovering Aggregation Functions} \label{sec: sub probab-rule-agg}
We will demonstrate in this section that the inference model leads to the different aggregation strategies depending on the assumed joint distribution when marginals are approximated with the rule confidences. Therefore we assume for the following derivations $p(\onerulRV_i{=}1) = \textit{conf}(\atomrule_i)$ for $i \in \indexSetAll$.
\subsubsection{Probabilistic Max-Aggregation}\label{sec: probab max aggregation}

Max-aggregation was introduced in the literature as a computational heuristic~\cite{amieplus2015}, it was further described as accounting for strong rule dependencies without providing a detailed treatment~\cite{meilicke2019anyburl}, or it was even described with assuming fact independence~\cite{svatovs2020strike}. We will now introduce the Fréchet-Hoeffding bound which will help us to achieve a formal derivation. It limits the possible association, expressed as correlation, of two random variables~\cite{joe1997multivariate}.
%completeely reworked..old version here
%In the recent literature \textcolor{red}{Max-Aggregation}  is either described as a computational heuristic~\cite{meilicke2019anyburl} \textcolor{red}{(you could maybe cite AMIE+ here they also describe max-aggregation)} or it is only vaguely described with the assumption that all rules are completely dependent~\cite{betz2022supervised}. If we would, however, assume complete dependence~\cite{lancaster1963correlation} between rules, e.g., $r_i$ takes the same value as $r_j$ with probability one, then clearly their marginals must be identical which is not the case for most of the rules.  
%More precisely, there exists a statistical property limiting the possible association, expressed as the correlation, of two Bernoulli variables with unequal probabilities \cite{iyengar1998multivariate}.
Let $p_i$ and $p_j$ be the marginal probabilities for two Bernoulli variables, then it holds for the correlation  $\rho_{ij}$ that $\rho_{ij} \leq U(i,j)$ where
\vspace{-0.2cm}
\begin{align}
    \resizebox{0.905\linewidth}{!}{$
     U(i,j) = \min \bigg\{ \bigg( \frac{p_i(1-p_j)}{p_j(1-p_i)}\bigg)^{1/2}, \bigg( \frac{p_j(1-p_i)}{p_i(1-p_j)}\bigg)^{1/2} \; \bigg\}. \label{eq: upper corr bound}
     $}
\end{align}
\begin{example}[cont]
Let $p_1=0.64\;$ and $\;p_2=0.44 \;$ then $U(1,2) \approx 0.66$. Whereas for $p_3=0.41$, $U(2,3) \approx 0.94$.  \label{ex: upper corr bound}
\end{example}
\noindent While the configuration of the marginals in Example \ref{ex: upper corr bound} allows for complex dependencies in regard to the joint distribution, they are not compatible with complete dependence as this would require unit correlation. Interestingly, equation~\eqref{eq: upper corr bound} suffices to specify a joint distribution $p(\vecallrules | \alltriples)$  such that the inference model from Section~\ref{sec: inference over the knowledge graph} performs Max-aggregation.

\begin{theorem} \label{theorem: max distribution}
    If for the correlation matrix $\Omega \in [-1,1]^{(N,N)}$ with entries $\rho_{ij}$ for all $i,j$ it holds that $\rho_{ij} = U(i,j)$ then a unique distribution for  $p(\vecallrules | \alltriples)$ is induced such that $p(t | \alltriples)=s^M(t).$
\end{theorem}

\noindent We will show the proof for the case where $k=2$ rules predicted the candidate here briefly and the general case can be found in the appendix. Let $p_{\bar{i}}=1-p_{i}$ and let, e.g., $p_{\bar{i}j}=p(R_i{=}0,R_j{=}1 |{\alltriples})$ and likewise for the remaining realisations. Further note for the correlation $\rho_{ij} = \frac{p_{ij}-p_ip_j}{\Tilde{\sigma}_i \Tilde{\sigma}_j}$ where $\Tilde{\sigma}$ is the respective standard deviation. \\

\noindent\textbf{Proof (k=2).} Following Propositions \eqref{prop: kvsN} and~\eqref{prop: oneIsTrue}, $p(t | \alltriples)$ is equivalent to querying the joint distribution marginally for $p(r_i + r_j \geq 1)$ assuming $\atomrule_i$ and $\atomrule_j$ predicted the target. We here assume the global distribution exists and is unique. It therefore suffices to show that
\begin{align*}
     \max{\{p_i, p_j\}} = p_{\bar{i}j} + p_{i\bar{j}} + p_{ij} \;.
\end{align*}
Assume w.l.o.g. that $p_i \geq p_j$. Then after plugging in $U(i,j)$ into $\rho_{ij}$ and solving for $p_{ij}$, we obtain  $p_{ij} = p_j$. However, by definition of the marginal it holds that $p_j= p_{ij} + p_{\bar{i}j}$ and therefore $p_{\bar{i}j}=0$. Then we have,
\begin{align*}
     \max{\{p_i, p_j\}} &=  \max{\{ p_{i\bar{j}}+ p_{ij}, \; p_{\bar{i}j} + p_{ij}\}}\\
     &=  \max{\{ p_{i\bar{j}}+ p_{ij}, \; p_{ij}\}}\\
     &=  p_{i\bar{j}} +  p_{ij}\\
     &=  p_{\bar{i}j} + p_{i\bar{j}} + p_{ij}. \tag*{\qed}
\end{align*}

\begin{example}[cont]
    For $p_{1} = 0.64$ and $p_2=0.44$ we obtain $p_{12} = 0.44$, $p_{\bar{1}2}=0$, and $p_{1\bar{2}}=p_1-p_2=0.2$, leading to $p(t | \alltriples) = 0 \cdot p_{\bar{1}\bar{2}} + 1 \cdot p_{12} + 1 \cdot p_{1\bar{2}} + 1 \cdot p_{\bar{1}2} = 0.64$.
\end{example}

\noindent We have specified a unique multivariate Bernoulli distribution $p(\vecallrules | \mathcal{G)}$ by simply defining a correlation matrix. Clearly setting the  $N^2$ values of the correlation matrix is in general not sufficient for defining a distribution that has $2^N$ parameters and also not every correlation matrix is admissible in the first place \cite{huber2019admissible}. %\textcolor{red}{Interestingly, this also makes it efficient to query this distribution for any probability although its use seems limited in other scenarios.}

\subsubsection{Noisy-or Aggregation} \label{sec: noisy-or}

To derive Noisy-or aggregation we have to make an assumption about the joint distribution that goes beyond pairwise interactions.
\begin{proposition} \label{prop: noisy-or-agg}
If the $N$ rules in $p(\randomvecallrules | \alltriples)$ are mutually independent then $p(t | \alltriples)=s^{NO}(t)$.
\end{proposition}
It is trivial to derive the Noisy-or product from the inference model under the independence assumption and the proof is shown in the appendix for completeness.

The independence assumption of Noisy-or aggregation reveals the connection of the model from section~\ref{sec: inference over the knowledge graph}  to ProbLog~\cite{de2007problog}. ProbLog assigns probabilities to logic programs and inference is performed by aggregating all programs that logically entail a query by assuming individual probabilities are independent. Two results are shown in the appendix that make the connection to the derivations here explicit. First, if the logical semantics of ProbLog would be substituted with one-step entailment than it would perform Noisy-or aggregation. Second if we setup a ProbLog program with the rules $\allatomrules$, the fact probability would be always equal or larger than the Noisy-or probability. Note that the computational complexity of reasoning, as discussed earlier, here also applies. Finally, aggregating all the predicting rules with the Noisy-or product might not optimal in the context of data-driven rule learning where millions of rules can be partially redundant, which will be shown in the experimental section.

%-----------------------------------------------------
\subsection{Mixing Assumptions}\label{sec:mixing}

Both of the aggregation approaches derived in Section~\ref{sec: sub probab-rule-agg} make strong assumptions in regard to the dependence structure of the joint distribution over the rules. Clearly this can lead to an overestimation or underestimation of the final probability when the assumptions fail. Intuitively, this gives rise to mixture distributions that make assumptions between mutual independence and maximal correlation. Along these lines, previous work proposes models that can express both approaches as their special cases. These models are expensive to use, however, as they learn a clustering of all rules~\cite{ott2021safran} or represent rules with latent embeddings~\cite{betz2022supervised}. We will now present a simple approach that is overlooked in the literature so far which likewise operates in between both assumptions.

\begin{definition}(Noisy-or top-h)
Let $\indexSetFired^* \subseteq \indexSetFired$ be the subset of indices for the $h$ predicting rules with the highest marginals. The Noisy-or top-h aggregation strategy calculates the final score according to $s(t)^{NO_h} = 1-\prod_{j \in \indexSetFired^*}(1-conf(\atomrule_j))$.
\end{definition}
The correlation assumption is revealed when considering that for decreasing $h$ the approach converges to Max-aggregation which is stated more compactly in the final proposition of this section.
\begin{proposition}
For the score calculated with noisy-or top-h we have that $s^{M}(t)\leq s^{NO_h}(t) \leq s^{NO}(t)$ where the equalities are achieved for $h=1$ and $h=k$, respectively.
\end{proposition}

The proposition immediately follows from the definitions of the approaches. Furthermore, instead of setting one value for $h$ we can exploit the mixture property more finegrained and set the value independently for relations and query-directions  which will be discussed in the next section.
%-----------------------------------------------------------------------------------------
\begin{table*}[t]
    \def\arraystretch{1.0}
    \centering
    \adjustbox{max width=1\textwidth}{%
    %\begin{tabular}{l@{\hskip 2em}l@{\hskip 2em}c@{\hskip 2em}c@{\hskip 2em}c@{\hskip 2em}c@{\hskip 2em}}
    \begin{tabular}{llc@{\hskip 0.75em}c@{\hskip 0.75em}c@{\hskip 1.5em}c@{\hskip 0.75em}c@{\hskip 0.75em}c@{\hskip 1.5em}c@{\hskip 0.75em}c@{\hskip 0.75em}c@{\hskip 1.5em}c@{\hskip 0.75em}c@{\hskip 0.75em}c@{\hskip 0.75em}}
    \toprule
    && \multicolumn{3}{c}{FB15k-237} & \multicolumn{3}{c}{WNRR} & \multicolumn{3}{c}{Codex-M} & \multicolumn{3}{c}{Yago3-10}  \\
    \midrule
    &Approach&  h@1 & h@10 & MRR & h@1 & h@10 & MRR & h@1 & h@10 &MRR & h@1 & h@10 &MRR\\
    \midrule
    \multirow{7}{*}{\begin{turn}{90}AnyBURL\end{turn}}
    & MAX  & 0.236 & 0.496 & 0.321 & 0.442 & 0.561 & 0.482 & 0.240 & 0.443 & 0.309 & 0.394 & 0.640 & 0.477 \\
    & MAX+  & 0.246 & 0.506 & 0.331 & 0.457 & 0.574 & 0.497 & 0.248 & 0.452 & 0.317 & 0.498 & 0.691 & 0.566 \\
    & NO  & 0.251 & 0.499 & 0.333 & 0.391 & 0.560 & 0.446 & 0.219 & 0.427 & 0.290 & 0.367 & 0.628 & 0.456 \\
    & NO top-5  & 0.260 & 0.524 & 0.347 & 0.458 & 0.578 & 0.499 & 0.243 & 0.461 & 0.317 & 0.486 & 0.697 & 0.560 \\
    & NO top-$h^*$  & 0.263 & 0.524 & 0.349 & 0.459 & 0.578 & 0.499 & 0.253 & 0.464 & 0.326 & 0.498 & 0.698 & 0.568 \\
    \cmidrule{2-14} 
    & SAFRAN  & 0.272 & 0.524 & 0.357 & 0.459 & 0.578 & 0.502 & 0.254 & 0.458 & 0.325 & 0.491 & 0.693 & 0.564 \\
    & SV  & 0.266 & 0.526 & 0.352 & 0.459 & 0.574 & 0.499 & 0.266 & 0.467 & 0.335 & - & - & - \\ 
    
     \midrule
    \multirow{5}{*}{\begin{turn}{90}AMIE3\end{turn}}& MAX  & 0.167 & 0.384 & 0.236 & 0.414 & 0.511 & 0.445 & 0.191 & 0.383 & 0.255 & 0.350 & 0.592 & 0.431 \\
    & MAX+  & 0.178 & 0.394 & 0.247 & 0.419 & 0.514 & 0.450 & 0.198 & 0.395 & 0.263 & 0.395 &  0.622 & 0.473 \\
    & NO  & 0.209 & 0.430 & 0.284 & 0.377 & 0.513 & 0.424 & 0.190 & 0.390 & 0.257 & 0.345 & 0.615 & 0.439 \\
    & NO top-5  & 0.199 & 0.425 & 0.273 & 0.380 & 0.513 & 0.426 & 0.197 & 0.401 & 0.266 & 0.360 & 0.622 & 0.452 \\
    & {NO top-$h^*$}  & 0.217 & 0.439 & 0.292 & 0.419 & 0.514 & 0.450 & 0.199 & 0.407 & 0.269 & 0.401 & 0.625 & 0.479 \\
     \midrule
    \toprule
    \end{tabular}
    }%
     \caption{Results for the joint filtered MRR and Hits@X with rules from AnyBURL or AMIE}.
     \label{table:joint-237-wnrr-codex}
\end{table*}
%----------------------------------------
\section{Experiments}\label{sec:experiments}
%---------------------------------------------------------------------
The goal of our experimental section is to analyse the predictive performance of the existing aggregation approaches, to evaluate how to efficiently exploit the overlooked Noisy-or top-$h$ approach, and to give a potential user an overview about the performance-speed trade-off regarding more complex approaches. We abstain from comparing against the general KGC literature which is not the focus of this work. The competitiveness of rule-based approaches is discussed in many works and we refer to the recent literature for a summary~\cite{rossi2020knowledge,sadeghian2019drum,meilicke2023largeAnyburl}.
%--------------------------------------------

\subsection{Experimental Settings}
We evaluate the aggregation techniques on the most common KGs from the KGC community. We use FB15k-237~\cite{toutanova2015observed}, WNRR~\cite{dettmers2018convolutionalConvE}, Codex-M~\cite{safavi-koutra-2020-codex}, and Yago3-10~\cite{dettmers2018convolutionalConvE}. The datasets are downloaded from the LibKGE library~\cite{broscheit2020libkge} and we use the same train, valid, testing splits as used throughout the literature as well as the exact same evaluation protocol~\cite{rossi2020knowledge} which is described in Section~\ref{sec: KGC}. 

We use AnyBURL~\cite{meilicke2019anyburl} and AMIE3~\cite{lajus2020fast} to mine the rulsets $\allatomrules$. For AnyBURL we use the same rulesets as used by Meilicke et al.~\yrcite{meilicke2021naive}. For AMIE3 we tried to find the best possible hyperparameter configuration regarding the results (see appendix).

We compare Max (MAX), Max+ (MAX+), Noisy-or (NO), and Noisy-or top-h aggregation (NO top-$h$). For Noisy-or top-h we investigate how one global value $h=5$ performs over all datasets and we additionally search for the best parameter on the validation set for the relations and query directions independently (NO top-$h^*$) as described in Section~\ref{sec:mixing}. For AnyBURL we search over the values $h \in \{1, 4 \ldots 10\}$ where for $h{=}1$ we use MAX+. For AMIE3 we additionally include $h{=}k$ as AMIE3 learned smaller rulesets and overall a smaller number of rules predict the query candidates. We also include the two works concerned with the aggregation problem, SAFRAN~\cite{ott2021safran} and the supervised sparse aggregator (SV) proposed by Betz et al. \yrcite{betz2022supervised}. We provide wall-clock times (Table~\ref{table:runtimes}) of the approaches for the larger datasets and the rulesets of AnyBURL. Further experimental details, the used server architecture, dataset statistics, and the overall number of learned rules can be found in the appendix of the work.

\subsection{Results}
Table~\ref{table:joint-237-wnrr-codex} shows performance results and Table~\ref{table:runtimes} shows runtimes for the rules from  AnyBURL. Despite the fact that the datasets are quite different NO top-5 performs surprisingly well and for the rules from AnyBURL it only falls short for the h@1 and MRR metrics for Yago3-10 compared to MAX+ while being faster on average and 1.6PP better on FB15k-237. In general we observe nevertheless that the best performing specification might be dataset specific, e.g., for the rules from AMIE3 NO performs best on FB15k-237, however, the results for these rulesets are significantly worse in general. A pragmatic approach is to simply learn the best value for $h$ on the validation set which, not surprisingly, performs always as good or better as the second best configuration although the improvement is sometimes marginal. 

Although SAFRAN and SV are superior on average in regard to performance they are significantly slower. For instance SAFRAN is outperformed on Codex-m by NO top-$h^*$ while running approximately 55 times longer and it is 0.8PP better on FB15k-237 where it runs more than 100 times longer. SV performs 0.3PP better on FB15k-237 while being 180 times slower and it performs 0.9PP better on Codex-M with a running time that is 13 times slower.

To conclude we observe that the aggregation method can have significant impact on the overall performance of the mined rulsets. Furthermore, when runtimes are a consideration factor a simple approach might be the preferred choice of aggregation.

%table with seconds
%\begin{table}[t]
%   \def\arraystretch{1.0}
%   \centering
%    \adjustbox{max width=0.45\textwidth}{%
%    %\begin{tabular}{l@{\hskip 2em}l@{\hskip 2em}c@{\hskip 2em}c@{\hskip 2em}c@{\hskip 2em}c@{\hskip 2em}}
%    \begin{tabular}{lc@{\hskip 1em}c@{\hskip 1em}c@{\hskip 1em}}
%    \toprule
%    Approach& {FB15k-237} & {Codex-M} & {Yago3-10}  \\
%    \midrule
%    MAX&63s&331s&247s\\
%    MAX+&185s&624s&249s\\
%    Noisy-or&321s&1498s&733s\\
%   Noisy-or top-5&88s&398s&258s\\
%    NO top-$h^*$&835s&1.27h&1.01\\
%    \midrule
%    SAFRAN &$\approx$24h&$\approx$72h&\textcolor{red}{X}\\
%    Supervised &$\approx$42h&$\approx$16.6h&-\\
%    \toprule
%    \end{tabular}
%    }%
%     \caption{\textcolor{red}{Aggregation runtimes in seconds (s) our hours (h) with rulesets from AnyBURL}.}
%     \label{table:runtimes}
%\end{table}

\begin{table}[h]
    \def\arraystretch{1.0}
    \centering
    \adjustbox{max width=0.5\textwidth}{%
    %\begin{tabular}{l@{\hskip 2em}l@{\hskip 2em}c@{\hskip 2em}c@{\hskip 2em}c@{\hskip 2em}c@{\hskip 2em}}
    \begin{tabular}{lc@{\hskip 1em}c@{\hskip 1em}c@{\hskip 1em}}
    \toprule
    Approach& {FB15k-237} & {Codex-M} & {Yago3-10}  \\
    \midrule
    MAX&1.1m&5.5m&4.1m\\
    MAX+&3.1m&10.4m&4.2m\\
    Noisy-or&5.4m&25.0m&12.2m\\
    Noisy-or top-5&1.5m&6.6m&4.3m\\
    NO top-$h^*$&13.9m&1.27h&1.01h\\
    \midrule
    SAFRAN &$\approx$24h&$\approx$72h&$>$72h\\
    SV &$\approx$42h&$\approx$16.5h&-\\
    \toprule
    \end{tabular}
    }%
     \caption{Runtimes in minutes (m) our hours (h) with rules from AnyBURL.}
     \label{table:runtimes}
\end{table}
%-----------------------------------------------------------------------------------------
\section{Conclusion}
We have shown that the problem of rule aggregation for KGC can be expressed with marginal inference over a joint distribution over the rules. We provided probabilistic interpretations for previously defined aggregation functions. Subsequently we proposed a baseline that is slightly superior over previous simple methods while being efficient and we found that more advanced models are expensive to use while only providing a small boost in regard to predictive performance. Future work might build on these foundations by finding suitable ways of modelling the joint distribution over the rules. For instance, rules could be grouped according to syntactic similarity, distributions might be estimated from more advanced statistics such as pairwise confidences or marginals could be approximated more rigorously. 

\cleardoublepage

\bibliography{main}

\begin{thebibliography}{48}
\providecommand{\natexlab}[1]{#1}
\providecommand{\url}[1]{\texttt{#1}}
\expandafter\ifx\csname urlstyle\endcsname\relax
  \providecommand{\doi}[1]{doi: #1}\else
  \providecommand{\doi}{doi: \begingroup \urlstyle{rm}\Url}\fi

\bibitem[Baier et~al.(2017)Baier, Ma, and Tresp]{Baier2017ImprovingVR}
Baier, S., Ma, Y., and Tresp, V.
\newblock Improving visual relationship detection using semantic modeling of
  scene descriptions.
\newblock In \emph{SEMWEB}, 2017.

\bibitem[Betz et~al.(2022{\natexlab{a}})Betz, Meilicke, and
  Stuckenschmidt]{betz2022adversarial}
Betz, P., Meilicke, C., and Stuckenschmidt, H.
\newblock Adversarial explanations for knowledge graph embedding models.
\newblock In \emph{Proceedings of the 31th International Joint Conference on
  Artificial Intelligence}, pp.\  2820--2826. Ijcai.org, 2022{\natexlab{a}}.

\bibitem[Betz et~al.(2022{\natexlab{b}})Betz, Meilicke, and
  Stuckenschmidt]{betz2022supervised}
Betz, P., Meilicke, C., and Stuckenschmidt, H.
\newblock Supervised knowledge aggregation for knowledge graph completion.
\newblock In \emph{Extended Semantic Web Conference}, pp.\  74--92. Springer,
  2022{\natexlab{b}}.

\bibitem[Broscheit et~al.(2020)Broscheit, Ruffinelli, Kochsiek, Betz, and
  Gemulla]{broscheit2020libkge}
Broscheit, S., Ruffinelli, D., Kochsiek, A., Betz, P., and Gemulla, R.
\newblock Libkge-a knowledge graph embedding library for reproducible research.
\newblock In \emph{Proceedings of the 2020 Conference on Empirical Methods in
  Natural Language Processing: System Demonstrations}, pp.\  165--174, 2020.

\bibitem[Chen et~al.(2016)Chen, Wang, and Goldberg]{ChenWG16}
Chen, Y., Wang, D.~Z., and Goldberg, S.
\newblock Scalekb: scalable learning and inference over large knowledge bases.
\newblock \emph{The VLDB Journal}, 25\penalty0 (6):\penalty0 893--918, 2016.

\bibitem[De~Raedt et~al.(2007)De~Raedt, Kimmig, and Toivonen]{de2007problog}
De~Raedt, L., Kimmig, A., and Toivonen, H.
\newblock Problog: A probabilistic prolog and its application in link
  discovery.
\newblock In \emph{Proceedings of the Twentieth International Joint Conference
  on Artificial Intelligence}, pp.\  2462--2467, 2007.

\bibitem[Dettmers et~al.(2018)Dettmers, Minervini, Stenetorp, and
  Riedel]{dettmers2018convolutionalConvE}
Dettmers, T., Minervini, P., Stenetorp, P., and Riedel, S.
\newblock Convolutional 2d knowledge graph embeddings.
\newblock In \emph{Proceedings of the AAAI Conference on Artificial
  Intelligence}, pp.\  1811--1818, 2018.

\bibitem[Evans \& Grefenstette(2018)Evans and Grefenstette]{evans2018learning}
Evans, R. and Grefenstette, E.
\newblock Learning explanatory rules from noisy data.
\newblock In \emph{Journal of Artificial Intelligence Research}, volume~61,
  pp.\  1--64, 2018.

\bibitem[Fan et~al.(2022)Fan, Fu, Jin, Lu, and Tian]{fan2022discovering}
Fan, W., Fu, W., Jin, R., Lu, P., and Tian, C.
\newblock Discovering association rules from big graphs.
\newblock \emph{Proceedings of the VLDB Endowment}, 15\penalty0 (7):\penalty0
  1479--1492, 2022.

\bibitem[Gal{\'a}rraga et~al.(2015)Gal{\'a}rraga, Teflioudi, Hose, and
  Suchanek]{amieplus2015}
Gal{\'a}rraga, L., Teflioudi, C., Hose, K., and Suchanek, F.~M.
\newblock Fast rule mining in ontological knowledge bases with {AMIE+}.
\newblock \emph{The VLDB Journal}, 24\penalty0 (6):\penalty0 707--730, 2015.

\bibitem[Gal{\'a}rraga et~al.(2013)Gal{\'a}rraga, Teflioudi, Hose, and
  Suchanek]{amie2013}
Gal{\'a}rraga, L.~A., Teflioudi, C., Hose, K., and Suchanek, F.
\newblock Amie: association rule mining under incomplete evidence in
  ontological knowledge bases.
\newblock In \emph{Proceedings of the 22nd international conference on World
  Wide Web}, pp.\  413--422. ACM, 2013.

\bibitem[Garc{\'{\i}}a{-}Dur{\'{a}}n \&
  Niepert(2018)Garc{\'{\i}}a{-}Dur{\'{a}}n and
  Niepert]{DBLP:conf/uai/Garcia-DuranN18}
Garc{\'{\i}}a{-}Dur{\'{a}}n, A. and Niepert, M.
\newblock Kblrn: End-to-end learning of knowledge base representations with
  latent, relational, and numerical features.
\newblock In Globerson, A. and Silva, R. (eds.), \emph{Proceedings of the
  Thirty-Fourth Conference on Uncertainty in Artificial Intelligence}, pp.\
  372--381. {AUAI} Press, 2018.

\bibitem[Guo et~al.(2016)Guo, Wang, Wang, Wang, and Guo]{guo2016jointly}
Guo, S., Wang, Q., Wang, L., Wang, B., and Guo, L.
\newblock Jointly embedding knowledge graphs and logical rules.
\newblock In \emph{Proceedings of the 2016 Conference on Empirical Methods in
  Natural Language Processing}, pp.\  192--202, 2016.

\bibitem[Guo et~al.(2018)Guo, Wang, Wang, Wang, and Guo]{guo2018knowledge}
Guo, S., Wang, Q., Wang, L., Wang, B., and Guo, L.
\newblock Knowledge graph embedding with iterative guidance from soft rules.
\newblock In \emph{Proceedings of the AAAI Conference on Artificial
  Intelligence}, 2018.

\bibitem[Huber \& Mari{\'c}(2019)Huber and Mari{\'c}]{huber2019admissible}
Huber, M. and Mari{\'c}, N.
\newblock Admissible bernoulli correlations.
\newblock \emph{Journal of Statistical Distributions and Applications},
  6\penalty0 (1):\penalty0 1--8, 2019.

\bibitem[Joe(1997)]{joe1997multivariate}
Joe, H.
\newblock \emph{Multivariate models and multivariate dependence concepts}.
\newblock CRC press, 1997.

\bibitem[Kersting \& De~Raedt(2001)Kersting and De~Raedt]{kersting2001towards}
Kersting, K. and De~Raedt, L.
\newblock Towards combining inductive logic programming with bayesian networks.
\newblock In \emph{Inductive Logic Programming: Proceedings of the 11th
  International Conference}, pp.\  118--131. Springer, 2001.

\bibitem[Kuzelka et~al.(2018)Kuzelka, Wang, Davis, and
  Schockaert]{kuzelka2018pac}
Kuzelka, O., Wang, Y., Davis, J., and Schockaert, S.
\newblock Pac-reasoning in relational domains.
\newblock In \emph{Proceedings of the Thirty-Fourth Conference on Uncertainty
  in Artificial Intelligence}, 2018.

\bibitem[Lajus et~al.(2020)Lajus, Gal{\'a}rraga, and Suchanek]{lajus2020fast}
Lajus, J., Gal{\'a}rraga, L., and Suchanek, F.
\newblock Fast and exact rule mining with amie 3.
\newblock In \emph{Proceedings of the Extended Semantic Web Conference}, pp.\
  36--52. Springer, 2020.

\bibitem[Liu et~al.(2021)Liu, Hildebrandt, Joblin, Ringsquandl, Raissouni, and
  Tresp]{hetioliu2021neural}
Liu, Y., Hildebrandt, M., Joblin, M., Ringsquandl, M., Raissouni, R., and
  Tresp, V.
\newblock Neural multi-hop reasoning with logical rules on biomedical knowledge
  graphs.
\newblock In \emph{European Semantic Web Conference}, pp.\  375--391. Springer,
  2021.

\bibitem[Meilicke et~al.(2018)Meilicke, Fink, Wang, Ruffinelli, Gemulla, and
  Stuckenschmidt]{rulen2018}
Meilicke, C., Fink, M., Wang, Y., Ruffinelli, D., Gemulla, R., and
  Stuckenschmidt, H.
\newblock Fine-grained evaluation of rule- and embedding-based systems for
  knowledge graph completion.
\newblock In \emph{Proceedings of the International Semantic Web Conference},
  pp.\  3--20. Springer, 2018.

\bibitem[Meilicke et~al.(2019)Meilicke, Chekol, Ruffinelli, and
  Stuckenschmidt]{meilicke2019anyburl}
Meilicke, C., Chekol, M.~W., Ruffinelli, D., and Stuckenschmidt, H.
\newblock Anytime bottom-up rule learning for knowledge graph completion.
\newblock In \emph{Proceedings of the Twenty-Eighth International Joint
  Conference on Artificial Intelligence}, pp.\  3137--3143. International Joint
  Conferences on Artificial Intelligence Organization, 2019.

\bibitem[Meilicke et~al.(2020)Meilicke, Chekol, Fink, and
  Stuckenschmidt]{meilicke2020reinforced}
Meilicke, C., Chekol, M.~W., Fink, M., and Stuckenschmidt, H.
\newblock Reinforced anytime bottom up rule learning for knowledge graph
  completion.
\newblock \emph{arXiv preprint arXiv:2004.04412}, 2020.

\bibitem[Meilicke et~al.(2021)Meilicke, Betz, and
  Stuckenschmidt]{meilicke2021naive}
Meilicke, C., Betz, P., and Stuckenschmidt, H.
\newblock Why a naive way to combine symbolic and latent knowledge base
  completion works surprisingly well.
\newblock In \emph{3rd Conference on Automated Knowledge Base Construction},
  2021.

\bibitem[Meilicke et~al.(2023)Meilicke, Chekol, Betz, Fink, and
  Stuckenschmidt]{meilicke2023largeAnyburl}
Meilicke, C., Chekol, M.~W., Betz, P., Fink, M., and Stuckenschmidt, H.
\newblock Anytime bottom-up rule learning for large scale knowledge graph
  completion.
\newblock \emph{The VLDB Journal—The International Journal on Very Large Data
  Bases}, 2023.

\bibitem[Minervini et~al.(2020{\natexlab{a}})Minervini, Bo{\v{s}}njak,
  Rockt{\"a}schel, Riedel, and Grefenstette]{minervini2020differentiableGNTP}
Minervini, P., Bo{\v{s}}njak, M., Rockt{\"a}schel, T., Riedel, S., and
  Grefenstette, E.
\newblock Differentiable reasoning on large knowledge bases and natural
  language.
\newblock In \emph{Proceedings of the AAAI Conference on Artificial
  Intelligence}, volume~34, pp.\  5182--5190, 2020{\natexlab{a}}.

\bibitem[Minervini et~al.(2020{\natexlab{b}})Minervini, Riedel, Stenetorp,
  Grefenstette, and Rockt{\"a}schel]{minervini2020learningCNTP}
Minervini, P., Riedel, S., Stenetorp, P., Grefenstette, E., and
  Rockt{\"a}schel, T.
\newblock Learning reasoning strategies in end-to-end differentiable proving.
\newblock In \emph{International Conference on Machine Learning}, pp.\
  6938--6949. PMLR, 2020{\natexlab{b}}.

\bibitem[Muggleton \& De~Raedt(1994)Muggleton and
  De~Raedt]{muggleton1994inductive}
Muggleton, S. and De~Raedt, L.
\newblock Inductive logic programming: Theory and methods.
\newblock \emph{The Journal of Logic Programming}, 19:\penalty0 629--679, 1994.

\bibitem[Muggleton~et al.(1996)]{muggleton1996stochastic}
Muggleton~et al., S.
\newblock Stochastic logic programs.
\newblock \emph{Advances in inductive logic programming}, 32:\penalty0
  254--264, 1996.

\bibitem[Noessner et~al.(2013)Noessner, Niepert, and
  Stuckenschmidt]{noessner2013rockit}
Noessner, J., Niepert, M., and Stuckenschmidt, H.
\newblock Rockit: Exploiting parallelism and symmetry for map inference in
  statistical relational models.
\newblock In \emph{Proceedings of the AAAI Conference on Artificial
  Intelligence}, volume~27, 2013.

\bibitem[Ortona et~al.(2018)Ortona, Meduri, and Papotti]{OrtonaMP18}
Ortona, S., Meduri, V.~V., and Papotti, P.
\newblock Robust discovery of positive and negative rules in knowledge bases.
\newblock In \emph{34th {IEEE} International Conference on Data Engineering},
  pp.\  1168--1179. {IEEE} Computer Society, 2018.

\bibitem[Ott et~al.(2021)Ott, Meilicke, and Samwald]{ott2021safran}
Ott, S., Meilicke, C., and Samwald, M.
\newblock {SAFRAN}: An interpretable, rule-based link prediction method
  outperforming embedding models.
\newblock In \emph{3rd Conference on Automated Knowledge Base Construction},
  2021.
\newblock URL \url{https://openreview.net/forum?id=jCt9S_3w_S9}.

\bibitem[Pearl(1988)]{pearl1988probabilistic}
Pearl, J.
\newblock \emph{Probabilistic reasoning in intelligent systems: networks of
  plausible inference}.
\newblock Morgan kaufmann, 1988.

\bibitem[Pellissier~Tanon et~al.(2017)Pellissier~Tanon, Stepanova, Razniewski,
  Mirza, and Weikum]{pellissier2017completeness}
Pellissier~Tanon, T., Stepanova, D., Razniewski, S., Mirza, P., and Weikum, G.
\newblock Completeness-aware rule learning from knowledge graphs.
\newblock In \emph{International Semantic Web Conference}, pp.\  507--525.
  Springer, 2017.

\bibitem[Qu et~al.(2021)Qu, Chen, Xhonneux, Bengio, and Tang]{qu2020rnnlogic}
Qu, M., Chen, J., Xhonneux, L.-P., Bengio, Y., and Tang, J.
\newblock Rnnlogic: Learning logic rules for reasoning on knowledge graphs.
\newblock \emph{Proceedings of the International Conference on Learning
  Representations}, 2021.

\bibitem[Richardson \& Domingos(2006)Richardson and
  Domingos]{richardson2006markov}
Richardson, M. and Domingos, P.
\newblock Markov logic networks.
\newblock In \emph{Machine learning}, volume~62, pp.\  107--136. Springer,
  2006.

\bibitem[Rockt{\"a}schel \& Riedel(2017)Rockt{\"a}schel and
  Riedel]{rocktaschel2017endNTP1}
Rockt{\"a}schel, T. and Riedel, S.
\newblock End-to-end differentiable proving.
\newblock In \emph{Advances in Neural Information Processing Systems 30: Annual
  Conference on Neural Information Processing Systems 2017}, pp.\  3788--3800,
  2017.

\bibitem[Rossi et~al.(2021)Rossi, Barbosa, Firmani, Matinata, and
  Merialdo]{rossi2020knowledge}
Rossi, A., Barbosa, D., Firmani, D., Matinata, A., and Merialdo, P.
\newblock Knowledge graph embedding for link prediction: A comparative
  analysis.
\newblock \emph{ACM Transactions on Knowledge Discovery from Data (TKDD)},
  15\penalty0 (2):\penalty0 1--49, 2021.

\bibitem[Ruffinelli et~al.(2020)Ruffinelli, Broscheit, and
  Gemulla]{ruffinelli2020dog}
Ruffinelli, D., Broscheit, S., and Gemulla, R.
\newblock You {CAN} teach an old dog new tricks! on training knowledge graph
  embeddings.
\newblock In \emph{8th International Conference on Learning Representations},
  2020.

\bibitem[Sadeghian et~al.(2019)Sadeghian, Armandpour, Ding, and
  Wang]{sadeghian2019drum}
Sadeghian, A., Armandpour, M., Ding, P., and Wang, D.~Z.
\newblock Drum: End-to-end differentiable rule mining on knowledge graphs.
\newblock In \emph{Advances in Neural Information Processing Systems}, pp.\
  15321--15331, 2019.

\bibitem[Safavi \& Koutra(2020)Safavi and Koutra]{safavi-koutra-2020-codex}
Safavi, T. and Koutra, D.
\newblock {C}o{DE}x: A {C}omprehensive {K}nowledge {G}raph {C}ompletion
  {B}enchmark.
\newblock In \emph{Proceedings of the 2020 Conference on Empirical Methods in
  Natural Language Processing}, pp.\  8328--8350. Association for Computational
  Linguistics, 2020.

\bibitem[Sato \& Kameya(1997)Sato and Kameya]{sato1997prism}
Sato, T. and Kameya, Y.
\newblock Prism: a language for symbolic-statistical modeling.
\newblock In \emph{Proceedings of the Fifteenth International Joint Conference
  on Artificial Intelligence}, volume~97, pp.\  1330--1339, 1997.

\bibitem[Svato{\v{s}} et~al.(2020)Svato{\v{s}}, Schockaert, Davis, and
  Ku{\v{z}}elka]{svatovs2020strike}
Svato{\v{s}}, M., Schockaert, S., Davis, J., and Ku{\v{z}}elka, O.
\newblock Strike: Rule-driven relational learning using stratified
  k-entailment.
\newblock In \emph{Proceedings of the European Conference of Artificial
  Intelligence}, 2020.

\bibitem[Tena~Cucala et~al.(2022)Tena~Cucala, Cuenca~Grau, and
  Motik]{faithful22Cucala}
Tena~Cucala, D.~J., Cuenca~Grau, B., and Motik, B.
\newblock {Faithful Approaches to Rule Learning}.
\newblock In \emph{{Proceedings of the 19th International Conference on
  Principles of Knowledge Representation and Reasoning}}, pp.\  484--493, 8
  2022.

\bibitem[Toutanova \& Chen(2015)Toutanova and Chen]{toutanova2015observed}
Toutanova, K. and Chen, D.
\newblock Observed versus latent features for knowledge base and text
  inference.
\newblock In \emph{Proceedings of the 3rd Workshop on Continuous Vector Space
  Models and their Compositionality}, pp.\  57--66, Beijing, China, July 2015.
  Association for Computational Linguistics.
\newblock \doi{10.18653/v1/W15-4007}.
\newblock URL \url{https://aclanthology.org/W15-4007}.

\bibitem[Wu et~al.(2022)Wu, Wang, Wang, and Shen]{wu2022learning}
Wu, H., Wang, Z., Wang, K., and Shen, Y.-D.
\newblock Learning typed rules over knowledge graphs.
\newblock In \emph{Proceedings of the International Conference on Principles of
  Knowledge Representation and Reasoning}, volume~19, pp.\  494--503, 2022.

\bibitem[Yang et~al.(2017)Yang, Yang, and Cohen]{yang2017differentiable}
Yang, F., Yang, Z., and Cohen, W.~W.
\newblock Differentiable learning of logical rules for knowledge base
  reasoning.
\newblock In \emph{Advances in Neural Information Processing Systems}, pp.\
  2319--2328, 2017.

\bibitem[Zupanc \& Davis(2018)Zupanc and Davis]{zupanc2018estimating}
Zupanc, K. and Davis, J.
\newblock Estimating rule quality for knowledge base completion with the
  relationship between coverage assumption.
\newblock In \emph{Proceedings of the 2018 World Wide Web Conference}, pp.\
  1073--1081, 2018.

\end{thebibliography}
\bibliographystyle{icml2023}
\appendix
\section{Proofs}

\noindent \textbf{Proof of Proposition \ref{prop: kvsN}.} We show that to perform inference $p(t | \alltriples)$ with the probabilistic model it is sufficient to perform marginal inference on the global distribution with respect to the rules that predicted the target candidate $t$.

\begin{align}
    p(t | \alltriples)&=\sum_{\vecallrules \in \{0, 1\}^N} p(t | \vecallrules, \alltriples)p(\vecallrules | \alltriples) \\
     &=\sum\limits_{\substack{\vecallrules \in \{0, 1\}^N \\ L(\vecallrules) \models_1 t}} p(\vecallrules | \alltriples) \label{eq:before split}
\end{align}

\noindent We will now split each $\vecallrules$ into vectors $\mathbf{r}^+=(\oneruleassignment_j)_{\substack{j \in \indexSetAll, \; c_j\ \models_1 t}}$, the rules that predicted the target, and $\mathbf{r}^-=(\oneruleassignment_i)_{\substack{i \in \indexSetAll, \; c_i\ \not\models_1 t}}$ the rules that did not predict the target. Let $\mathbf{r}^+||\mathbf{r}^- = \vecallrules$, where $||$ denotes vector concatenation. Now we can write equation \eqref{eq:before split} as

%In the last expression the $\vecallrules's$ still contain all the rules of the knowledge base. The rules that did not predict the query, however, do not influence $L(\vecallrules) \models_1 t_1$. We will now split each $\vecallrules$ into vectors $\vecallfiredrules$, the rules that predicted the query, and $\mathbf{r}^{\not \models_1}$ representing the rules that did not predict the query, i.e., $L(\mathbf{r}^{\not \models_1}) \not \models_1 t $. Equation \eqref{eq:before split} becomes
\begin{align}
     &\sum\limits_{\substack{ \mathbf{r}^+ || \mathbf{r}^{-} \in \{0, 1\}^N \\ L(\vecallfiredrules^{+} || \mathbf{r}^-) \models_1 t}} p(\mathbf{r}^+, \mathbf{r}^- | \alltriples) \\
    &= \sum\limits_{\substack{ \mathbf{r}^+ \in \{0, 1\}^k \\ L(\mathbf{r}^+) \models_1 t}} 
    \; \; \sum\limits_{\substack{ \mathbf{r}^{-} \in \{0, 1\}^{N-k} \\  L(\mathbf{r}^-) \not\models_1 t}} p(\mathbf{r}^+, \mathbf{r}^{-} | \alltriples). \label{eq: continue}
\end{align}
Observe that the inner sum contains all possible values of $\mathbf{r}^{-}$ as $ L(\mathbf{r}^-) \not\models_1 t$ does not put a constraint on $\mathbf{r}^-$. Continuing from equation \eqref{eq: continue} we can therefore simply apply reverse marginalization,

\begin{align*}
    &= \sum\limits_{\substack{ \vecallfiredrules^+ \in \{0, 1\}^k \\ L(\vecallfiredrules^+) \models_1 t}} 
     p(\vecallfiredrules^+ | \alltriples) \\
     &=\sum_{\vecallfiredrules\in \{0, 1\}^k}  p(t |\vecallfiredrules, \alltriples) p(\vecallfiredrules | \alltriples) . \tag*{\qed}
\end{align*}
  \\
 \\
%---------------------------------------------------------------------------
\noindent \textbf{Proof of Theorem~\ref{theorem: max distribution}}.
This proof is a generalisation of the binary case from section~\ref{sec: probab max aggregation}. We first show that under maximal correlation only very specific realisations of $\randomvecallrules$ have non-zero probability if the distribution exists. Once this is established we show the existence and uniqueness of $p(\randomvecallrules | \alltriples)$ and finally we derive the max-aggregation score from the marginal inference that at least one of the predicting rules is true.
As previously we assume that out of $N$ rules $k$ rules predicted the query triple $t$ and that the rule marginals are given. After we have specified $p(\randomvecallrules | \alltriples)$, by Proposition~\ref{prop: kvsN} and~\ref{prop: oneIsTrue}, we have to show that
\begin{align}
   1 - p(\mathbf{R}=\mathbf{0} | \alltriples) = \max \big\{p(\oneFiredrulRV_i=1) \; | \; \atomrule_i \models_1 t \textrm{ and } i \in \indexSetFired  \big\}, \label{eq: main-result-max}
\end{align}
where $\mathbf{R}$ is the random vector for the $k$ rules that predicted the query triple. We assume throughout the derivations the $N$ variables $\{\onerulRV_1, ..., \onerulRV_N \}$ are ordered by $\indexSetAll$ (and likewise for $\indexSetFired$) such that $\onerulRV_1$ is the rule with the highest marginal.

First we pick two rules represented by $\onerulRV_i$, $\onerulRV_j$ with $p_i \geq p_j$ and $i,j \in \indexSetAll$. The correlation is defined as
\begin{align}
    \rho_{ij} = \frac{p_{ij} - p_ip_j}{\Tilde{\sigma_i} \Tilde{\sigma}_j} \label{eq:corr},
\end{align}
Where $p_{ij} = p(\onerulRV_i{=}1, \onerulRV_j{=}1 |\alltriples)$. We now assume $\rho_{ij}{=}U(i,j)$ for every pair $i,j$ from $\indexSetAll$. We plug in the upper bound~\eqref{eq: upper corr bound} into~\eqref{eq:corr} and solve for $p_{ij}$ which leads to 
\begin{align}
   p_{ij} &= p_j\label{eq: marginal1},
\end{align}
i.e., we have the following equality
\begin{align*}
   \sum_{\Tilde{\vecallfiredrules}_{-ij} \in \{0, 1\}^{N-2}}p(\onerulRV_i{=}1, \onerulRV_j{=}1, \Tilde{\vecallfiredrules}_{-ij} | \alltriples) = 
\end{align*}
\begin{align}
   =\sum_{\Tilde{\vecallfiredrules}_{-j} \in \{0, 1\}^{N-1}}p(\onerulRV_j{=}1, \Tilde{\vecallfiredrules}_{-j} | \alltriples) \label{eq: marginal2},
\end{align}

\noindent where $\Tilde{\vecallfiredrules}_{-j}= (\oneruleassignment_i)_{i \in \indexSetAll \setminus j}$ is a vector of realisations with the $j$'th component dropped from $\Tilde{\vecallfiredrules}$ and 
equivalently $\vecallrules_{-ij}= (\oneruleassignment_s)_{s \in \indexSetAll  \setminus \{i,j\}}$.
Each addend in the left hand side is contained in the right hand side, subtracting the left hand side from both sides of~\eqref{eq: marginal2} yields a zero probability constraint:
\begin{align}
   0=\sum_{\vecallfiredrules_{-ij} \in \{0, 1\}^{N-2}}p(\onerulRV_i=0, \onerulRV_j=1, \vecallfiredrules_{-ij} | \alltriples) \label{eq: marginal-zero-constraint}.
\end{align}

%The maximal correlation assumption enforces a zero probability to every configuration in which the rule with the smaller marginal is true and the rule with the higher marginal is false. For specifying $p(\Tilde{\mathbf{R}} | \alltriples)$ we require $2^N$ parameters. Already from equation~\eqref{eq: marginal-zero-constraint} alone we have $2^{N-2}$ zero-constraints on these parameters. While each ordered rule pair induces another $2^{N-2}$ set of constraints most of them are overlapping. 

\noindent We are, in fact, interested in all the realisations that may be different from zero after considering the constraints imposed by all possible rule pairs. From equation \eqref{eq: marginal-zero-constraint} it follows that $p(\Tilde{\mathbf{R}} = \vecallrules | \alltriples)$ is not affected by the zero-constraint if for $\vecallrules$
\begin{align}
    (\oneruleassignment_s = 0) \implies (\oneruleassignment_t = 0) \;\; \forall \; t > s , \label{eq:admissable rule}
\end{align}
for $s,t \in \indexSetAll$. Note that each assignment $\vecallrules \in \{0, 1\}^N$ which satisfies~\eqref{eq:admissable rule} is associated with a unique number of components (rules) that are one.

%by the implication in~\eqref{eq:admissable rule} each assignment $\vecallrules \in \{0, 1\}^N$ does not have a probability of zero is associated with a unique number of components (rules) that are one.

Our goal is to specify the parameters of $p(\randomvecallrules | \alltriples)$ that are non-zero. From~\eqref{eq:admissable rule} we can observe that there are only $N+1$ of these parameters left and we will therefore introduce $N+1$ variables. Let $m \in \{0,...,N\}$ and let $z_m$ denote the probability for the assignment vector that has $m$ ones and satisfies~\eqref{eq:admissable rule} which we write as $\vecallrules^{(m)}$, i.e., $z_m = p({\vecallrules^{(m)}} | \alltriples)$. In fact, $\vecallrules^{(m)} \in \{0, 1\}^N$ holds ones from the first component until the $m$'th component and zeros starting from the $m+1$'th component. It is easy to verify that now it holds for the $N$ marginals with $i \in \indexSetAll$ that $p_i = \sum_{s=i}^{N}z_s$ and additionally we use the probability constraint $p_0 =  \sum_{s=0}^{N}z_s = 1$. With these expressions we can set up an equation system
\begin{align}
    \mathbf{A} \mathbf{z} = \mathbf{p} \label{eq:system},
\end{align}
where $\mathbf{z}$ is the variable vector with dimensionality $N+1$, $\mathbf{A}$ is an upper triangular coefficient matrix with all non-zero entries being one, and $\mathbf{p}$ is the vector of marginals and the probability constraint at the first entry. Given that $\mathbf{A}$ is invertible we established uniqueness and we established existence as the solution $\mathbf{z} = \mathbf{A}^{-1}\mathbf{p}$ satisfies the probability constraint $\sum_{s=0}^{N}z_s = 1$ while all $z_m$ are between 0 and 1. 

We will now derive the main result from~\eqref{eq: main-result-max}. Plugging in the expressions for the marginals in the right-hand side of~\eqref{eq: main-result-max} yields
\begin{align}
    \max \big\{\sum_{s=i}^{N}z_s \; \big| \; \atomrule_i \models_1 t \textrm{ and } i \in \indexSetFired  \big\} &=  \label{eq:the end}
       \sum_{s=s^*}^{N}z_s \;,
\end{align}
\noindent  where $s^* = \min \{i \;|\; \atomrule_i \models_1 t \; \textrm{and} \; i \in \indexSetFired\}$ corresponds to the index for the rule with the highest marginal under the predicting rules. For $1-p(\mathbf{R}{=}\mathbf{0} | \alltriples)$ we have to sum up all probabilities of realisations where at least one of the predicting rules is one. Clearly this includes all realisations where $r_{s^*}$ is one which holds by construction of the $z_m$'s for every term in the sum on the right hand side of equation~\eqref{eq:the end}. Now given that the remaining probabilities are zero we have that  $\sum_{s=s^*}^{N}z_s = 1-p(\mathbf{R}=\mathbf{0} | \alltriples). $\qed \\
\\
%--------------------------------------------------------------------------------------------
\noindent\textbf{Proof of Proposition~\ref{prop: noisy-or-agg}.} We directly start with the result from Proposition~\ref{prop: kvsN}.

%extended version
%\begin{align}
%    p(t | \alltriples) &= \sum_{\mathbf{r} \in \{0,1\}^k}  p(t |  \mathbf{r}, \alltriples)p(\mathbf{r} | \alltriples)  \\
%    &=  0 \cdot p(\mathbf{r}=\mathbf{0}) +  \sum_{\mathbf{r}\neq \mathbf{0} \in \{0,1\}^k} p(\mathbf{r} | \alltriples)  \\
%      &=  1  -  p(\mathbf{r}=\mathbf{0} | \alltriples)  \\
%    &= 1-\prod_{i=1}^{k} \big(1-p(r_i)\big)
%\end{align}

\begin{align*}
    p(t | \alltriples) &= \sum_{\vecallfiredrules \in \{0,1\}^k}  p(t |  \vecallfiredrules, \alltriples)p(\vecallfiredrules | \alltriples)  \\
    &=  0 \cdot p(\mathbf{R}=\mathbf{0} | \alltriples) +  \sum_{\vecallfiredrules \neq \mathbf{0} \in \{0,1\}^k} p(\vecallfiredrules | \alltriples)  \\
     &= 1 - p(\mathbf{R}=\mathbf{0} |\alltriples)\\
    &= 1-\prod_{j \in \indexSetFired} \big(1-p(r_j)\big) \tag*{\qed}
\end{align*}
%------------------------------------------------
\\

%------------------------------------------------
\section{Evaluation Metrics}
Let $\mathcal{G}_e$ be the test KG. For the target test fact $q(c_1,c_2)$ let $\mathbf{rk}(c_1|c_2,q)$ be the ranking position of the target in a filtered ranking of candidate facts for the tail query $q(c_1,?)$ and likewise let $\mathbf{rk}(c_2|q,c_1)$ denote the filtered ranking position for the head query. MRR and Hits@X are defined as: 

{\begin{align*}
    MRR &= \frac{1}{2 |\mathcal{G}_e|} \sum_{q(c_1,c_2) \in \mathcal{G}_e} \bigg( \frac{1}{\mathbf{rk}(c_1|c_2,q)} + \frac{1}{\mathbf{rk}(c_2|q,c_1)} \bigg),
\end{align*}

\begin{align*}
     hits@X &= \frac{1}{2 |\mathcal{G}_e|} \sum_{q(c_1,c_2) \in \mathcal{G}_e} \bigg(  \mathtt{1}\big\{\mathbf{rk}(c_1|c_2,q) \leq X \big\}
     &+ \\
     &+\mathtt{1} \big\{\mathbf{rk}(c_2|q,c_1)\leq X \big\} \bigg).
\end{align*}

\section{Rule Aggregation and Reasoning with Problog}\label{sec: reasoning and aggregation}
%---------------------------------------------------------
The probability for a logic program $\probloglogicprogram$ given a ProbLog program $T$ is defined as  
\begin{align}
    p( \probloglogicprogram | T) = \prod_{x_i \in \probloglogicprogram} p(x_i) \prod_{x_j \not\in \probloglogicprogram} (1-p(x_j)), \label{eq: problog program prob}
\end{align}
\noindent where $T$ is a collection of definite clauses with assigned probabilities. In the scope of this work when using the ProbLog notation we can set $T^* = \{p_i: \atomrule_i \;|\; i \in \indexSetAll \} \cup \{1:t' \:|\; t' \in \alltriples\}$ to obtain the ProbLog program representing the rules and the facts. If we, for instance, let ProbLog only perform one-step entailment we obtain the following result.

\begin{proposition} \label{prop: problog one step}
    For the probability $p(t | T^*)$ calculated with ProbLog under a one-step entailment regime it holds that $p(t | T^*)=s^{NO}(t)$.
\end{proposition}

\textbf{Proof of Proposition~\ref{prop: problog one step}}. The query probability given the ProbLog program $T^*$, as defined above, is calculated as
\begin{align}
    p(t | T^*) = \sum_{\vecallrules \in \{0,1\}^N} p\big(t | L(\vecallrules)\big)p\big(L(\vecallrules)|T^*\big) \;, \label{eq: problog inference}
\end{align}
where $p(t | L(\vecallrules))$ is set to equation~\eqref{eq:logical part} by requirement of the proposition and $p(L(\vecallrules) | T)$ can be interpreted as the probability of a logic program when treating rules as logical clauses and facts as ground atoms. Note that $L(\vecallrules) =  \{\atomrule_i  \;| \; \oneruleassignment_i = 1 \; \textit{}{and} \; i \in \indexSetAll \} \cup \alltriples$. Plugging in equation~\eqref{eq: problog program prob} into equation~\eqref{eq: problog inference}  and rearranging leads to:
\resizebox{\linewidth}{!}{
\begin{minipage}{\linewidth}
\begin{align*}
    &\sum_{\vecallrules \in \{0,1\}^N} p(t | L(\vecallrules))\prod_{x_i \in L(\vecallrules)} p(x_i) \prod_{x_j \not\in L(\vecallrules)} (1-p(x_j))\\
    &=\sum_{\mathbf{\vecallrules} \in \{0,1\}^N} p(t | L(\vecallrules))\prod_{\atomrule_i \in L(\vecallrules)} p(\oneruleassignment_i) \prod_{\atomrule_j \not\in L(\vecallrules)} (1-p(\oneruleassignment_j))\prod_{t \in \alltriples } p(t) \\
    &=\sum_{\mathbf{\vecallrules} \in \{0,1\}^N} p(t | L(\vecallrules))\prod_{\atomrule_i \in L(\vecallrules)} p(\oneruleassignment_i) \prod_{\atomrule_j \not\in L(\vecallrules)} (1-p(\oneruleassignment_j)) \cdot 1 \\
    &=  \sum_{\mathbf{\vecallrules} \in \{0,1\}^N} p(t | L(\vecallrules)) p(\vecallrules | \alltriples) \;.
\end{align*}
\end{minipage}
}
The factorization of the logic program implies mutual independence and therefore applying Propositions~\ref{prop: kvsN} and~\ref{prop: noisy-or-agg} leads to the Noisy-or product over the rules that predicted $t$. \qed

The next theorem shows the behaviour when using the full ProbLog algorithm for rule aggregation.

\begin{theorem} \label{theorem: full problog vs noisy-or}
For the query probability $p(t | T^*)$ calculated by ProbLog it holds that  $p(t | T^*) \geq s^{NO}(t)$.
\end{theorem}

\noindent We first sketch the proof here which is straightforward when using Propositions~\ref{prop: problog one step} and~\ref{prop: kvsN}. The details are given below. ProbLog sums the probabilities of all programs that entail the target fact. This includes 1) the  programs that entail and one-step entail the query and 2) the programs that entail but not one-step entail the query which is clearly larger or equal than only aggregating 1) as done in Noisy-or aggregation.

\subsection*{Proof of Theorem~\ref{theorem: full problog vs noisy-or}}
As before let $\vecallrules \in \{0,1\}^N$ be a vector of realisations. We will now label different assignment vectors according to their logical properties. Let $\vecallrules^{(e)} \in \{0, 1\}^N$ be an assignment vector such that $L(\vecallrules^{(e)})$ entails but not one-step entails the query and let $\vecallrules^{(o)} \in \{0, 1\}^N$ be the corresponding vector where $L(\vecallrules^{(o)})$ entails and one-step entails the query. Let $T^*$ denote the ProbLog program as defined above. For the query probability under ProbLog we have
\begin{align}
    p(t | T^*) &= \sum_{\vecallrules \in \{0,1\}^N} \hat{p}\big(t | L(\vecallrules)\big)p\big(L(\vecallrules)|T^*\big) \label{eq: start-problog},
\end{align}
where
\begin{align*}
     \hat{p}(t | L(\vecallrules))= \left\{ \begin{array}{l}
    1, \textrm{ if $L(\vecallrules) \models t$   } \ \\
    0, \textrm{ else}.
  \end{array}\right.
\end{align*}
When we plug in $\hat{p}(t | L(\vecallrules))$  then~\eqref{eq: start-problog} becomes
\begin{align*}
   p(t | T^*)&=\sum\limits_{\substack{\vecallrules \in \{0,1\}^N \\ L(\vecallrules) \models t}} p(L(\vecallrules)|T^*) \\
     &= \sum\limits_{\substack{\vecallrules^{(e)} \in \{0,1\}^N \\ L(\mathbf{r}^{(e)}) \models t \\  L(\mathbf{r}^{(e)}) \not\models_1 t}} p(L(\vecallrules^{(e)})|T^*)+\\ &+\sum\limits_{\substack{\vecallrules^{(o)} \in \{0,1\}^N  \\  L(\vecallrules^{(o)}) \models_1 t } } p(L(\vecallrules^{(o)})|T^*)\\
      &\geq \sum\limits_{\substack{\vecallrules^{(o)} \in \{0,1\}^N  \\  L(\vecallrules^{(o)}) \models_1 t } } p(L(\vecallrules^{(o)})|T^*)\\
      &= \sum_{\vecallrules \in \{0,1\}^N} p\big(t | L(\vecallrules) \big)p\big(L(\vecallrules)|T^*\big)
\end{align*}
where it follows from Proposition~\ref{prop: problog one step} that the last expression is the Noisy-or product under the program factorization of ProbLog for $p\big(L(\vecallrules)|T^*\big)$ \qed

\section{Rule Aggregation and MLNs}
For an MLN query answering $p(\triple | \alltriples)$ as in the main text can be performed by marginal inference given some evidence which is in our case the KG. Nevertheless we start with a simple example with unary predicates. For the MLN definitions we refer to the original publication. Consider the two formulae $smokes(X) \rightarrow cancer(X)$ and $ill(X) \rightarrow cancer(X)$, assume they have some confidence $conf_1$ and $conf_2$. Now let the evidence be $e=\{smokes(karl), \; ill(karl)\}$ and we do not have any other constant terms. Note that there are $2^3$ possible worlds and we seek to calculate $p(cancer(karl){=}1| e)$, i.e., the probability that Karl has cancer. We abuse notation slightly for brevity. In particular we write $p(\{ a_1, a_2\})$ for the probability of a world where $a_i$ is a true atom and all the remaining possible atoms are false. And we write $p(a_i{=}1 | e)$ for the marginal probability that an atom is true given the evidence.

For instantiating a Markov Network, as each formula has one possible grounding, we have two binary features $f_1, f_2$. A feature is one if in a given world the respective formula is satisfied. Assigning the confidences as weights is not useful because of the exponential formulation, therefore we set the feature weights to the log odds $w_i=\ln \frac{conf_i}{1-conf_1}$ for $i \in \{1,2\}$. Now assume $conf_1=conf_2=0.9$. For $p(\{cancer(karl)\}| e)$ we have
\begin{align*}
    p(c(k)=1| e) &= \frac{p\big(\big\{c(k), i(k), sm(k)\big\}\big)}{p\big(sm(k)=1, i(k)=1\big)}
\end{align*}
Note that the denominator is a marginal that sums over all worlds where $smokes(karl)$ and $ill(karl)$ is true, therefore
\begin{align*}
    p(c(k)=1| e) &= \frac{p\big(\big\{c(k), i(k), sm(k)\big\}\big)}{p\big(\big\{c(k), i(k), sm(k)\big\}\big) + p\big(\big\{ i(k), sm(k)\big\}\big) }\\
    &=\frac{\exp{(w_1 + w_2)}Z^{-1}}{\exp{(w_1 + w_2)}Z^{-1} + \exp{(0)}Z^{-1}}\\
    &=\mathtt{sigmoid}(w_1+w_2)
\end{align*}
Plugging in the values for the weights results in 0.8581 which we also exactly recover when using the MLN solver Rockit~\cite{noessner2013rockit} under exact marginal inference. Now consider the rules,
\begin{align*}
c_1{:} \; married(X,Y) &\leftarrow engaged(X,Y) \\
c_2{:} \; married(X,Y) &\leftarrow commonChild(X,Y)  \\
c_3{:} \; married(X,Y) &\leftarrow inLove(X,Y) 
\end{align*}
Assume the hypothetical confidences $conf_1=0.8$, $conf_2=0.7$, $conf_3=0.5$. Further assume the evidence KG $\alltriples=\{inLove(a, b), commonChild(a, b), engaged(a, b) \}$ where $a,b$ are constants. We want to calculate $p(t=1 | \alltriples)$ where $t=married(a,b)$.  If we use ProbLog with the confidences we obtain $p(t | \alltriples)=0.97$ which is here the same as Noisy-or aggregation. Max-aggregation leads to $p(t | \alltriples)=0.8$ and using Max-group aggregation would be in between depending on the grouping. When using the log-odds for the MLN as above we obtain with Rockit  $p(t | \alltriples)=0.9032$ which is again the sigmoid function applied to the the sum of the log-odds.

Note that in these two examples the only form of reasoning in regard to the query fact is one-step entailment due to the simplicity of the examples. This can be emulated for MLN's also for more complicated examples with the common solvers by defining the rule bodies as observed predicates. Note that additionally the log odds would be weighted with each rule grounding regarding the rules that predicted the query. Nevertheless the resulting aggregation baseline would not perform well in the context of KGC as, for instance, a rule with confidence of 0.4 and many groundings would lead to a small value for the final probability. However, note that, as we mentioned in the main text, a MLN is a more general framework not based on one-step entailment and expressing the aggregation problem requires to make several non trivial decisions.

\section{Experimental Details}
%------------------------------------------

We show dataset statistics and the overall number of learned rules in Table~\ref{table:data statistics} and~\ref{table: numLearnedRules}. Further experimental details are given in the following subsections.

\begin{table}[h]
    \def\arraystretch{1.0}
    \centering
    \adjustbox{max width=0.4\textwidth}{%
    %\begin{tabular}{l@{\hskip 2em}l@{\hskip 2em}c@{\hskip 2em}c@{\hskip 2em}c@{\hskip 2em}c@{\hskip 2em}}
   \begin{tabular}{c|cc|ccc} 
&&& \multicolumn{3}{ c }{ Num facts $|\mathcal{\alltriples}|$} \\
Dataset & $|\mathcal{E}|$ & $|\mathcal{P}|$ &  Train & Valid & Test \\
\hline
FB15k-237 &  14 505  &  237 & 272 115  & 17 535 & 20 466  \\
WNRR & 40 559   &  11 & 86 835  & 3 034 & 3 134 \\
Codex-M  & 17 050 & 51 & 185 584  & 10 310 & 10 311 \\
Yago3-10 & 123 182 & 37 & 1 079 040 & 5 000 & 5 000 \\
\end{tabular}
    }%
     \caption{Dataset summary statistics}
     \label{table:data statistics}
\end{table}

\begin{table}[h]
    \def\arraystretch{1.0}
    \centering
    \adjustbox{max width=0.30\textwidth}{%
    %\begin{tabular}{l@{\hskip 2em}l@{\hskip 2em}c@{\hskip 2em}c@{\hskip 2em}c@{\hskip 2em}c@{\hskip 2em}}
   \begin{tabular}{c|c|c} 
Dataset & AMIE & AnyBURL  \\
\hline
FB15k-237 & 983 546  &  5 084 903  \\
WNRR & 3426   &  97 329  \\
Codex-M  & 179 898 & 7 409 385  \\
Yago3-10 & 900 951 & 6 692 784\\
\end{tabular}
    }%
     \caption{Number of rules learned}
     \label{table: numLearnedRules}
\end{table}

\subsection{Rule learning}
For AnyBURL we use the rulesets provided in previous work~\cite{meilicke2021naive} except for Yago3-10 where we learn with the default parameters for 3600 seconds with the AnyBURL-JUNO version. For AMIE3 under the default parameters only a few rules are learned so we adjusted the parameters. When including rules with constants and longer rules the approach does not terminate within a day therefore we use one execution with rules of length one including constants and one execution with longer rules without constants and subsequently the rulsets are merged. We show below the program execution with constants and the second execution with longer rules.  
\begin{lstlisting}[breaklines]
   java -jar amie.jar train.txt -const -bias default -minhc 0.0 -minc 0.001 -minpca 0.001 -maxad 2
\end{lstlisting}

\begin{lstlisting}[breaklines]
    java -jar amie.jar train.txt -bias default -minhc 0.0 -minc 0.001 -minpca 0.001 -maxad 4 -pm support -mins 2
\end{lstlisting}

\subsection{Implementation and Evaluation Details}
Rule-based KGC uses commonly a parameter $topX$ that denotes how many candidate facts $q(c_1, c^*)$ for the query  $q(c_1,  ?)$ (likewise in head direction) should be predicted and ranked. For all our experiments we set $topX{=}200$ but we did not notice significant differences to using 100. When two candidates are assigned to the same probability we use random tie handling. The aggregation functions are implemented under the AnyBURL-JUNO codebase. For MAX+ we use the existing implementation. For Noisy-or top-$h$ we start predicting with the rule with the highest marginal continuing with the second highest rule and we stop when for $topX$ candidates each one was predicted by at least $h$ rules. For Noisy-or we set $h=k$ and for MAX we set $h=1$.

\subsection{Experiment Details and Execution}
Our experiments are run on a CPU server with 768 GB RAM and two Intel(R) Xeon(R) CPU E5-2640 v4 @ 2.40GHz cores with 40 logical cores each. For the standard experiments we use the AnyBURL-JUNO codebase with 30 threads. For the Noisy-or top-$h^*$ approach we spawn individual processes with 15 threads each for every value of $h$. The runtimes for the experiments are wall-clock times, i.e., we measure the time before and after the execution on each dataset. For SAFRAN we obtained runtime estimates from the authors for FB15k-237 and Codex-m and we run the approach (architecture as above) on Yago3-10 where it did not terminate within 3 days. Runtimes for SV are also obtained from the authors. Results for SAFRAN are obtained from the authors where on FB15k-237 SAFRAN is run in accordance to a newer AnyBURL version that does not exploit the characteristic that connected entities in the training set cannot form a fact in the test set. See Meilicke et al.~\yrcite{meilicke2020reinforced} for a discussion. The results for SV are from the respective publication.
\end{document}